# The FF Planning System: Fast Plan Generation Through Heuristic Search

**Jörg Hoffmann**                                    HOFFMANN@INFORMATIK.UNI-FREIBURG.DE
**Bernhard Nebel**                                    NEBEL@INFORMATIK.UNI-FREIBURG.DE
*Georges-Köhler-Allee, Geb. 52,*
*79110 Freiburg, Germany*

## Abstract

We describe and evaluate the algorithmic techniques that are used in the FF planning system. Like the HSP system, FF relies on forward state space search, using a heuristic that estimates goal distances by ignoring delete lists. Unlike HSP's heuristic, our method does not assume facts to be independent. We introduce a novel search strategy that combines hill-climbing with systematic search, and we show how other powerful heuristic information can be extracted and used to prune the search space. FF was the most successful automatic planner at the recent AIPS-2000 planning competition. We review the results of the competition, give data for other benchmark domains, and investigate the reasons for the runtime performance of FF compared to HSP.

## 1. Introduction

Over the last few years we have seen a significant increase of the efficiency of planning systems. This increase is mainly due to three new approaches in plan generation.

The first approach was developed by Blum and Furst (1995, 1997). In their seminal paper on the GRAPHPLAN system (Blum & Furst, 1995), they described a new plan generation technique based on *planning graphs*, which was much faster than any other technique known at this time. Their paper started a whole series of research efforts that refined this approach by making it even more efficient (Fox & Long, 1998; Kambhampati, Parker, & Lambrecht, 1997) and by extending it to cope with more expressive planning languages (Koehler, Nebel, Hoffmann, & Dimopoulos, 1997; Gazen & Knoblock, 1997; Anderson, Smith, & Weld, 1998; Nebel, 2000).

The second approach is the *planning as satisfiability* method, which translates planning to propositional satisfiability (Kautz & Selman, 1996). In particular there is the hope that advances in the state of the art of propositional reasoning systems carry directly over to planning systems relying on this technology. In fact, Kautz and Selman (1999) predicted that research on planning methods will become superfluous because the state of the art in propositional reasoning systems will advance much faster than in planning systems.

A third new approach is *heuristic-search planning* as proposed by Bonet and Geffner (1998, 1999). In this approach a heuristic function is derived from the specification of the planning instance and used for guiding the search through the state space. As demonstrated by the system FF (short for Fast-Forward) at the planning competition at AIPS-2000, this approach proved to be competitive. In fact, FF outperformed all the other fully automatic systems and was nominated *Group A Distinguished Performance Planning System* at the competition.

In HSP (Bonet & Geffner, 1998), goal distances are estimated by approximating solution length to a relaxation of the planning task (Bonet, Loerincs, & Geffner, 1997). While FF uses the same relaxation for deriving its heuristics, it differs from HSP in a number of important details. Its base heuristic technique can be seen as an application of GRAPHPLAN to the relaxation. This yields goal distance estimates that, in difference to HSP's estimates, do not rely on an independence assumption. FF uses a different search technique than HSP, namely an *enforced form* of hill-climbing, combining local and systematic search. Finally, it employs a powerful pruning technique that selects a set of promising successors to each search node, and another pruning technique that cuts out branches where it appears that some goal has been achieved too early. Both techniques are obtained as a side effect of the base heuristic method.

Concerning the research strategy that FF is based on, we remark the following. A lot of classical planning approaches, like partial-order planning (McAllester & Rosenblitt, 1991) or planning graph analysis (Blum & Furst, 1997), are generic problem solving methods, developed following some theoretical concept, and tested on examples from the literature afterwards. In our approach, exploring the idea of heuristic search, there is no such clear distinction between development and testing. The search strategy, as well as the pruning techniques, are generic methods that have been motivated by observing examples. Also, design decisions were made on the basis of careful experimentation. This introduces into the system a bias towards the examples used for testing during development. We were testing our algorithms on a range of domains often used in the planning literature. Throughout the paper, we will refer to domains that are frequently used in the literature, and to tasks from such domains, as *benchmarks*. In the development phase, we used benchmark examples from the *Assembly*, *Blocksworld*, *Grid*, *Gripper*, *Logistics*, *Mystery*, *Mprime*, and *Tireworld* domains. When describing our algorithms in the paper, we indicate the points where those testing examples played a role for design decision making.

Planning is known to be PSPACE-complete even in its simplest form (Bylander, 1994). Thus, in the general case, there is no efficient algorithmic method. It is therefore worthwhile to look for algorithms that are efficient at least on restricted subclasses. To some extent, this idea has been pursued by posing severe syntactical restrictions to the planning task specifications (Bylander, 1994). Our approach is complementary to this. Examining the existing benchmarks, one finds that they, indeed, do not exploit the full expressivity of the underlying planning formalism. Though they do not fulfill any obvious rigid syntactical restrictions, almost none of them is particularly hard. In almost all of the existing benchmark domains, a non-optimal plan can, in principle, be generated in polynomial time. Using the benchmarks for inspiration during development, we have been able to come up with a heuristic method that is not provably efficient, but does work well empirically on a large class of planning tasks. This class includes almost all of the current planning benchmarks. Intuitively, the algorithms exploit the simple structure underlying these tasks. Our ongoing work is concerned with finding a formal characterization of that "simple" structure, and thereby formalizing the class of planning tasks that FF works well on.

Section 2 gives a schematic view on FF's system architecture, and Section 3 introduces our notational conventions for STRIPS domains. Sections 4 to 6 describe the base heuristic technique, search algorithm, and pruning methods, respectively. Section 7 shows how the algorithms are extended to deal with ADL domains. System performance is evaluated in Section 8, demonstrating that FF generates solutions extremely fast in a large range of planning benchmark domains. In order to illustrate our intuitions on the kind of structure that FF can exploit successfully, the section also gives examples of domains where the method is less appropriate. Finally, to clarify the performance differences between FF and HSP, the section describes a number of experiments we made in order to estimate which of the new algorithmic techniques is most useful. We show connections to related work at the points in the text where they apply, and overview other connections in Section 9. Section 10 outlines our current avenue of research.

## 2. System Architecture

To give the reader an overview of FF's system architecture, Figure 1 shows how FF's most fundamental techniques are arranged.

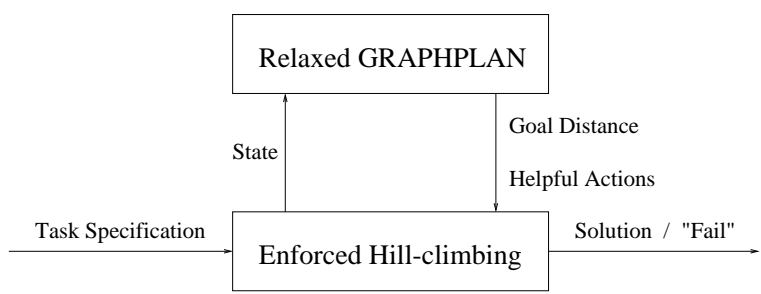

Figure 1: FF's base system architecture.

The fundamental heuristic technique in FF is relaxed GRAPHPLAN, which we will describe in Section 4. The technique gets called on every search state by *enforced hill-climbing*, our search algorithm. This is a forward searching engine, to be described in Section 5. Given a state, relaxed GRAPHPLAN informs the search with a goal distance estimate, and additionally with a set of promising successors for the state, the *helpful actions*, to be described in Section 6. Upon termination, enforced hill-climbing either outputs a solution plan, or reports that it has failed.

On top of the base architecture shown in Figure 1, we have integrated a few optimizations to cope with special cases that arose during testing:

- If a planning task contains states from which the goal is unreachable (dead ends, defined in Section 5.2), then enforced hill-climbing can fail to find a solution. In that case, a complete heuristic search engine is invoked to solve the task from scratch.

- In the presence of goal orderings, enforced hill-climbing sometimes wastes a lot of time achieving goals that need to be cared for later on. Two techniques trying to avoid this are integrated:

  - *Added goal deletion*, introduced in Section 6.2, cuts out branches where some goal has apparently been achieved too early.

  - The *goal agenda* technique, adapted from work by Jana Koehler (1998), feeds the goals to the planner in an order determined as a pre-process (Section 6.2.2).

## 3. Notational Conventions

For introducing FF's basic techniques, we consider simple STRIPS planning tasks, as were introduced by Fikes and Nilsson (1971). Our notations are as follows.

**Definition 1 (State)** *A state $S$ is a finite set of logical atoms.*

We assume that all operator schemata are grounded, i.e., we only talk about *actions*.

**Definition 2 (Strips Action)** *A STRIPS action $o$ is a triple*

$$o = (pre(o), add(o), del(o))$$

*where $pre(o)$ are the* preconditions *of $o$, $add(o)$ is the* add list *of $o$ and $del(o)$ is the* delete list *of the action, each being a set of atoms. For an atom $f \in add(o)$, we say that $o$ achieves $f$. The result of applying a single STRIPS action to a state is defined as follows:*

$$Result(S, \langle o \rangle) = \begin{cases} (S \cup add(o)) \setminus del(o) & pre(o) \subseteq S \\ undefined & otherwise \end{cases}$$

*In the first case, where $pre(o) \subseteq S$, the action is said to be* applicable *in $S$. The result of applying a sequence of more than one action to a state is recursively defined as*

$$Result(S, \langle o_1, \ldots, o_n \rangle) = Result(Result(S, \langle o_1, \ldots, o_{n-1} \rangle), \langle o_n \rangle).$$

**Definition 3 (Planning Task)** *A* planning task *$\mathcal{P} = (\mathcal{O}, \mathcal{I}, \mathcal{G})$ is a triple where $\mathcal{O}$ is the set of actions, and $\mathcal{I}$ (the initial state) and $\mathcal{G}$ (the goals) are sets of atoms.*

Our heuristic method is based on *relaxed planning tasks*, which are defined as follows.

**Definition 4 (Relaxed Planning Task)** *Given a planning task $\mathcal{P} = (\mathcal{O}, \mathcal{I}, \mathcal{G})$. The relaxation $\mathcal{P}'$ of $\mathcal{P}$ is defined as $\mathcal{P}' = (\mathcal{O}', \mathcal{I}, \mathcal{G})$, with*

$$\mathcal{O}' = \{(pre(o), add(o), \emptyset) \mid (pre(o), add(o), del(o)) \in \mathcal{O}\}$$

In words, one obtains the relaxed planning task by ignoring the delete lists of all actions. Plans are simple sequences of actions in our framework.

**Definition 5 (Plan)** *Given a planning task $\mathcal{P} = (\mathcal{O}, \mathcal{I}, \mathcal{G})$. A* plan *is a sequence $P = \langle o_1, \ldots, o_n \rangle$ of actions in $\mathcal{O}$ that solves the task, i.e., for which $\mathcal{G} \subseteq Result(\mathcal{I}, P)$ holds. An action sequence is called a* relaxed plan *to $\mathcal{P}$, iff it solves the relaxation $\mathcal{P}'$ of $\mathcal{P}$.*

## 4. GRAPHPLAN as a Heuristic Estimator

In this section, we introduce the base heuristic method used in FF. It is derived by applying GRAPHPLAN to relaxed planning tasks. The resulting goal distance estimates do not, like HSP's estimates, rely on an independence assumption. We prove that the heuristic computation is polynomial, give some notions on how distance estimates can be kept cautious, and describe how the method can be implemented efficiently.

Consider the heuristic method that is used in HSP (Bonet & Geffner, 1998). Given a planning task $\mathcal{P} = (\mathcal{O}, \mathcal{I}, \mathcal{G})$, HSP estimates for each state $S$ that is reached in a forward search the solution length of the task $\mathcal{P}'_S = (\mathcal{O}', S, \mathcal{G})$, i.e., the length of a relaxed plan that achieves the goals starting out from $S$. As computing the optimal solution length to $\mathcal{P}'_S$—which would make an admissible heuristic—is NP-hard (Bylander, 1994), the HSP estimate is a rough approximation based on computing the following weight values.

$$weight_S(f) := \begin{cases} 0 & \text{if } f \in S \\ i & \text{if } [min_{o \in \mathcal{O}, f \in \text{add}(o)} \sum_{p \in \text{pre}(o)} weight_S(p)] = i - 1 \\ \infty & \text{otherwise} \end{cases} \quad (1)$$

HSP assumes facts to be achieved independently in the sense that the weight of a set of facts—an action's preconditions—is estimated as the *sum* of the individual weights. The state's heuristic estimate is

$$h(S) := weight_S(\mathcal{G}) = \sum_{g \in \mathcal{G}} weight_S(g) \quad (2)$$

Assuming facts to be achieved independently, this heuristic ignores positive interactions that can occur. Consider the following short example planning task, where the initial state is empty, the goals are $\{G_1, G_2\}$, and there are the following three actions:

| name | | (pre, | add, | del) |
|------|---|-------|------|------|
| $\mathbf{op}G_1$ | $=$ | $(\{P\},$ | $\{G_1\},$ | $\emptyset)$ |
| $\mathbf{op}G_2$ | $=$ | $(\{P\},$ | $\{G_2\},$ | $\emptyset)$ |
| $\mathbf{op}P$ | $=$ | $(\emptyset,$ | $\{P\},$ | $\emptyset)$ |

HSP's weight value computation results in $P$ having weight one, and each goal having weight two. Assuming facts to be achieved independently, the distance of the initial state to a goal state is therefore estimated to four. Obviously, however, the task is solvable in only three steps, as $\mathbf{op}G_1$ and $\mathbf{op}G_2$ share the precondition $P$. In order to take account of such positive interactions, our idea is to start GRAPHPLAN on the tasks $(\mathcal{O}', S, \mathcal{G})$, and extract an explicit solution, i.e., a relaxed plan. One can then use this plan for heuristic evaluation. We will see in the next section that this approach is feasible: GRAPHPLAN can be proven to solve relaxed tasks in polynomial time.

### 4.1 Planning Graphs for Relaxed Tasks

Let us examine how GRAPHPLAN behaves when it is started on a planning task that does not contain any delete lists. We briefly review the basic notations of the GRAPHPLAN algorithm (Blum & Furst, 1997).

A planning graph is a directed, layered graph that contains two kinds of nodes: *fact nodes* and *action nodes*. The layers alternate between fact and action layers, where one fact and action layer together make up a *time step*. In the first time step, number 0, we have the fact layer corresponding to the initial state and the action layer corresponding to all actions that are applicable in the initial state. In each subsequent time step $i$, we have the layer of all facts that can possibly be made true in $i$ time steps, and the layer of all actions that are possibly applicable given those facts.

One crucial thing that GRAPHPLAN does when building the planning graph is the inference of *mutual exclusion* relations. A pair of actions $(o, o')$ at time step 0 is marked mutually exclusive, if $o$ and $o'$ *interfere*, i.e., if one action deletes a precondition or an add effect of the other. A pair of facts $(f, f')$ at a time step $i > 0$ is marked mutually exclusive, if each action at level $i - 1$ that achieves $f$ is exclusive of each action at level $i - 1$ that achieves $f'$. A pair of actions $(o, o')$ at a time step $i > 0$ is marked mutually exclusive, if the actions interfere, or if they have *competing needs*, i.e., if some precondition of $o$ is exclusive of some precondition of $o'$. The planning graph of a relaxed task does not contain any exclusion relations at all.

**Proposition 1** *Let $\mathcal{P}' = (\mathcal{O}', \mathcal{I}, \mathcal{G})$ be a relaxed STRIPS task. Started on $\mathcal{P}'$, GRAPH-PLAN will not mark any pair of facts or actions as mutually exclusive.*

**Proof:** The Proposition is easily proven by induction over the depth of the planning graph.

*Base case: time step* 0. Only interfering actions are marked mutual exclusive at time step 0. As there are no delete effects, no pair of actions interferes.

*Inductive case: time step $i \rightarrow$ time step $i + 1$.* Per induction hypothesis, the facts are not exclusive as their achievers one time step ahead are not. From this it follows that no pair of actions has competing needs. They do not interfere either. □

When started on a planning task, GRAPHPLAN extends the planning graph layer by layer until a fact layer is reached that contains all goal facts, and in which no two goal facts are marked exclusive.[1] Starting from that layer, a recursive backward search algorithm is invoked. To find a plan for a set of facts at layer $i > 0$, initialize the set of selected actions at layer $i - 1$ to the empty set. Then, for each fact, consider all achieving actions at layer $i - 1$ one after the other and select the first one that is not exclusive of any action that has already been selected. If there exists such an action, proceed with the next fact. If not, backtrack to the last fact and try to achieve it with a different action. If an achieving action has been selected for each fact, then collect the preconditions of all these actions to make up a new set of facts one time step earlier. Succeed when fact layer 0—the initial state—is reached, where no achieving actions need to be selected. On relaxed tasks, no backtracking occurs in GRAPHPLAN's search algorithm.

**Proposition 2** *Let $\mathcal{P}' = (\mathcal{O}', \mathcal{I}, \mathcal{G})$ be a relaxed STRIPS task. Started on $\mathcal{P}'$, GRAPH-PLAN will never backtrack.*

**Proof:** Backtracking only occurs if all achievers for a fact $f$ are exclusive of some already selected action. With Proposition 1, we know that no exclusions exist, and thus, that this does not happen. Also, if $f$ is in graph layer $i$, then there is at least one achiever in layer $i - 1$ supporting it. $\qquad\square$

---

1. If no such fact layer can be reached, then the task is proven to be unsolvable (Blum & Furst, 1997).

While the above argumentation is sufficient for showing Proposition 2, it does not tell us much about what is actually going on when one starts GRAPHPLAN on a task without delete lists. What happens is this. Given the task is solvable, the planning graph gets extended until some fact layer is reached that contains all the goals. Then the recursive search starts by selecting achievers for the goals at this level. The first attempt succeeds, and new goals are set up one time step earlier. Again, the first selection of achievers succeeds, and so forth, until the initial state is reached. Thus, search performs only a single sweep over the graph, starting from the top layer going down to the initial layer, and collects a relaxed plan on its way. In particular, the procedure takes only polynomial time in the size of the task.

**Theorem 1** *Let $\mathcal{P}' = (\mathcal{O}', \mathcal{I}, \mathcal{G})$ be a solvable relaxed STRIPS task, where the length of the longest add list of any action is $l$. Then GRAPHPLAN will find a solution to $\mathcal{P}'$ in time polynomial in $l$, $|\mathcal{O}'|$ and $|\mathcal{I}|$.*

**Proof:** Building the planning graph is polynomial in $l$, $|\mathcal{O}'|$, $|\mathcal{I}|$ and $t$, where $t$ is the number of time steps built (Blum & Furst, 1997). Now, in our case the total number $|\mathcal{O}'|$ of actions is an upper limit to the number of time steps. This is just because after this number of time steps has been built, all actions appear at some layer in the graph. Otherwise, there is a layer $i$ where no new action comes in, i.e., action layer $i - 1$ is identical to action layer $i$. As the task is solvable, this implies that all goals are contained in fact layer $i$, which would have made the process stop right away. Similarly, action layer $|\mathcal{O}'|$ would be identical to action layer $|\mathcal{O}'| - 1$, implying termination. The graph building phase is thus polynomial in $l$, $|\mathcal{O}'|$ and $|\mathcal{I}|$.

Concerning the plan extraction phase: With Proposition 2, search traverses the graph from top to bottom, collecting a set of achieving actions at each layer. Selecting achievers for a set of facts is $O(l * |\mathcal{O}'| + |\mathcal{I}|)$: A set of facts has at most size $l * |\mathcal{O}'| + |\mathcal{I}|$, the maximal number of distinct facts in the graph. An achieving action can be found to each fact in constant time using the planning graph. As the number of layers to be looked at is $O(|\mathcal{O}'|)$, search is polynomial in the desired parameters. $\qquad\square$

Starting GRAPHPLAN on a solvable search state task $(\mathcal{O}', S, \mathcal{G})$ yields—in polynomial time, with Theorem 1—a relaxed solution $\langle O_0, \ldots, O_{m-1} \rangle$, where each $O_i$ is the set of actions selected in parallel at time step $i$, and $m$ is the number of the first fact layer containing all goals. As we are interested in an estimation of *sequential* solution length, we define our heuristic as follows.

$$h(S) := \sum_{i=0,\ldots,m-1} |O_i| \qquad (3)$$

The estimation values obtained this way are, on our testing examples, usually lower than HSP's estimates (Equations 1 and 2), as extracting a plan takes account of positive interactions between facts. Consider again the short example from the beginning of this section, empty initial state, two goals $\{G_1, G_2\}$, and three actions:

| name | | (pre, | add, | del) |
|------|---|-------|------|------|
| $\mathbf{op}G_1$ | = | $(\{P\},$ | $\{G_1\},$ | $\emptyset)$ |
| $\mathbf{op}G_2$ | = | $(\{P\},$ | $\{G_2\},$ | $\emptyset)$ |
| $\mathbf{op}P$ | = | $(\emptyset,$ | $\{P\},$ | $\emptyset)$ |

Starting GRAPHPLAN on the initial state, the goals are contained in fact layer two, causing selection of $\mathbf{op}G_1$ and $\mathbf{op}G_2$ in action layer one. This yields the new goal $P$ at fact layer one, which is achieved with $\mathbf{op}P$. The resulting plan is $\langle\{\ \mathbf{op}P\ \}, \{\ \mathbf{op}G_1, \mathbf{op}G_2\ \}\rangle$, giving us the correct goal distance estimate three, as distinct from HSP's estimate four.

## 4.2 Solution Length Optimization

We use GRAPHPLAN's heuristic estimates, Equation 3, in a greedy strategy, to be introduced in Section 5.1, that does not take its decisions back once it has made them. From our experience with running this strategy on our testing examples, this works best when distance estimates are cautious, i.e., as low as possible. As already said, an *optimal* sequential solution can not be synthesized efficiently. What one *can* do is apply some techniques to make GRAPHPLAN return as short solutions as possible. Below, we describe some ways of doing that. The first technique is a built-in feature of GRAPHPLAN and ensures a minimality criterion for the relaxed plan. The two other techniques are heuristic optimizations.

### 4.2.1 NOOPs-first

The original GRAPHPLAN algorithm makes extensive use of so-called *NOOP*s. These are dummy actions that simply propagate facts from one fact layer to the next. For each fact $f$ that gets inserted into some fact layer, a NOOP corresponding to that fact is inserted into the action layer at the same time step. This NOOP has no other effect than adding $f$, and no other precondition than $f$. When performing backward search, the NOOPs are considered just like any other achiever, i.e., one way of making a fact true at time $i > 0$ is to simply keep it true from time $i - 1$.

In GRAPHPLAN, the implementation uses as a default the *NOOPs-first* heuristic, i.e., if there is a NOOP present for achieving a fact $f$, then this NOOP is considered first, before the planner tries selecting other "real" actions that achieve $f$. On relaxed tasks, the NOOPs-first heuristic ensures a minimality criterion for the returned plan as follows.

**Proposition 3** *Let* $(\mathcal{O}', \mathcal{I}, \mathcal{G})$ *be a relaxed STRIPS task, which is solvable. Using the NOOPs-first strategy, the plan that GRAPHPLAN returns will contain each action at most once.*

**Proof:** Let us assume the opposite, i.e., one action $o$ occurs twice in the plan $\langle O_0, \ldots, O_{m-1}\rangle$ that GRAPHPLAN finds. We have $o \in O_i$ and $o \in O_j$ for some layers $i, j$ with $i < j$.

Now, the action $o$ has been selected at layer $j$ to achieve some fact $f$ at layer $j + 1$. As the algorithm is using the NOOPs-first strategy, this implies that there is no NOOP for fact $f$ contained in action layer $j$: otherwise, the NOOP—not action $o$—would have been selected for achieving $f$.

In contradiction to this, action layer $j$ does indeed contain a NOOP for fact $f$. This is because action $o$ already appears in action layer $i < j$. As $f$ gets added by $o$, it appears in fact layer $i + 1 \leq j$. Therefore, a NOOP for $f$ is inserted in action layer $i + 1 \leq j$, and, in turn, will be inserted into each action layer $i' \geq i + 1$. □

### 4.2.2 Difficulty Heuristic

With the above argumentation, if we can achieve a fact by using a NOOP, we should do that. The question is, which achiever should we choose when no NOOP is available? It is certainly a good idea to select an achiever whose preconditions seem to be "easy". From the graph building phase, we can obtain a simple measure for the difficulty of an action's preconditions as follows.

$$\text{difficulty}(o) := \sum_{p \in \text{pre}(o)} min\{i \mid p \text{ is member of the fact layer at time step } i\} \quad (4)$$

The difficulty of each action can be set when it is first inserted into the graph. During plan extraction, facing a fact for which no NOOP is available, we then simply select an achieving action with minimal difficulty. This heuristic works well in situations where there are several ways to achieve one fact, but some ways need less effort than others.

### 4.2.3 Action Set Linearization

Assume GRAPHPLAN has settled for a parallel set $O_i$ of achievers at a time step $i$, i.e., achieving actions have been selected for all goals at time step $i + 1$. As we are only interested in sequential solution length, we still have a choice on how to linearize the actions. Some linearizations can lead to shorter plans than others. If an action $o \in O_i$ adds a precondition $p$ of another action $o' \in O_i$, then we do not need to include $p$ in the new set of facts to be achieved one time step earlier, given that we restrict ourselves to execute $o$ before $o'$. The question now is, how do we find a linearization of the actions that minimizes our new fact set? The corresponding decision problem is NP-complete.

**Definition 6** *Let* OPTIMAL ACTION LINEARIZATION *denote the following problem.*

Given a set $\mathcal{O}$ of relaxed STRIPS actions and a positive integer $K$. Is there a one-to-one function $f : \mathcal{O} \mapsto \{1, 2, \ldots, |\mathcal{O}|\}$ such that the number of unsatisfied preconditions when executing the sequence $\langle f^{-1}(1), \ldots, f^{-1}(|\mathcal{O}|)\rangle$ is at most $K$ ?

**Theorem 2** *Deciding* OPTIMAL ACTION LINEARIZATION *is NP-complete.*

**Proof:** Membership is obvious. Hardness is proven by transformation from DIRECTED OPTIMAL LINEAR ARRANGEMENT (Even & Shiloach, 1975). Given a directed graph $G = (V, A)$ and a positive integer $K$, the question is, does there exists a one-to-one function $f : V \mapsto \{1, 2, \ldots, |V|\}$ such that $f(u) < f(v)$ whenever $(u, v) \in A$ and such that $\sum_{(u,v) \in A}(f(v) - f(u)) \leq K$ ?

To a given directed graph, we define a set of actions as follows. For each node $w$ in the graph, we define an action in our set $\mathcal{O}$. For simplicity of presentation, we identify the actions with their corresponding nodes. To begin with, we set $\text{pre}(w) = \text{add}(w) = \emptyset$ for all $w \in V$. Then, for each edge $(u, v) \in A$, we create new logical facts $P_w^{(u,v)}$ and $R_w^{(u,v)}$ for $w \in V$.

Using these new logical facts, we now adjust all precondition and add lists to express the constraint that is given by the edge $(u, v)$. Say action $u$ is ordered before action $v$ in a linearization. We need to simulate the difference between the positions of $v$ and $u$. To do this, we define our actions in a way such that the bigger this difference is, the more unsatisfied preconditions there are when executing the linearization. First, we "punish" all actions that are ordered before $v$, by giving them an unsatisfied precondition.

$$\text{pre}(w) := \text{pre}(w) \cup P_w^{(u,v)} \text{ for } w \in V, \text{ add}(v) := \text{add}(v) \cup \{P_w^{(u,v)} \mid w \in V\}$$

With this definition, the actions $w$ ordered before $v$—and $v$ itself—will have the unsatisfied precondition $P_w^{(u,v)}$, while those ordered after will get this precondition added by $v$. Thus, the number of unsatisfied preconditions we get here is exactly $f(v)$.

Secondly, we "give a reward" to each action that is ordered *before* $u$. We simply do this by letting those actions add a precondition of $u$, which would otherwise go unsatisfied.

$$\text{add}(w) := \text{add}(w) \cup R_w^{(u,v)} \text{ for } w \in V, \text{ pre}(u) := \text{pre}(u) \cup \{R_w^{(u,v)} \mid w \in V\}$$

That way, we will have exactly $|V| - (f(u) - 1)$ unsatisfied preconditions, namely the $R_w^{(u,v)}$ facts for all actions except those that are ordered before $u$.

Summing up the number of unsatisfied preconditions we get for a linearization $f$, we arrive at

$$\sum_{(u,v) \in A} (f(v) + |V| - (f(u) - 1)) = \sum_{(u,v) \in A} (f(v) - f(u)) + |A| * (|V| + 1)$$

We thus define our new positive integer $K' := K + |A| * (|V| + 1)$.

Finally, we make sure that actions $u$ get ordered before actions $v$ for $(u, v) \in A$. We do this by inserting new logical "safety" facts $S_1^{(u,v)}, \ldots, S_{K'+1}^{(u,v)}$ into $v$'s precondition- and $u$'s add list.

$$\text{pre}(v) := \text{pre}(v) \cup \{S_1^{(u,v)}, \ldots, S_{K'+1}^{(u,v)}\}, \text{ add}(u) := \text{add}(u) \cup \{S_1^{(u,v)}, \ldots, S_{K'+1}^{(u,v)}\}$$

Altogether, a linearization $f$ of our actions leads to at most $K'$ unsatisfied preconditions if and only if $f$ satisfies the requirements for a directed optimal linear arrangement. Obviously, the action set and $K'$ can be computed in polynomial time. $\qquad \square$

Our sole purpose with linearizing an action set in a certain order is to achieve a smaller number of unsatisfied preconditions, which, in turn, might lead to a shorter relaxed solution.[2] Thus, we are certainly not willing to pay the price that finding an optimal linearization of the actions is likely to cost, according to Theorem 2. There are a few methods how one can approximate such a linearization, like introducing an ordering constraint $o < o'$ for each action $o$ that adds a precondition of another action $o'$, and trying to linearize the actions such that many of these constraints are met. During our experimentations, we found that parallel actions adding each other's preconditions occur so rarely in our testing tasks that even approximating is not worth the effort. We thus simply linearize all actions in the order they get selected, causing almost no computational overhead at all.

---

2. It should be noted here that using optimal action linearizations at each time step does *not* guarantee the resulting relaxed solution to be optimal, which would give us an admissible heuristic.

### 4.3 Efficient Implementation

We have implemented our own version of GRAPHPLAN, highly optimized for solving relaxed planning tasks. It exploits the fact that the planning graph of a relaxed task does not contain any exclusion relations (Proposition 1). Our implementation is also highly optimized for repeatedly solving planning tasks which all share the same set of actions—the tasks $\mathcal{P}'_S = (\mathcal{O}', S, \mathcal{G})$ as described at the beginning of this section.

Planning task specifications usually contain some operator schemata, and a set of constants. Instantiating the schemata with the constants yields the actions to the task. Our system instantiates all operator schemata in a way such that all, and only, reachable actions are built. Reachability of an action here means that, when successively applying operators to the initial state, all of the action's preconditions appear eventually. We then build what we call the connectivity graph. This graph consists of two layers, one containing all (reachable) actions, and the other all (reachable) facts. From each action, there are pointers to all preconditions, add effects and delete effects. All of FF's computations are efficiently implemented using this graph structure. For the subsequently described implementation of relaxed GRAPHPLAN, we only need the information about preconditions and add effects.

As a relaxed planning graph does not contain any exclusion relations, the only information one needs to represent it are what we call the *layer memberships*, i.e., for each fact or action, the number of the first layer at which it appears in the graph. Called on an intermediate task $\mathcal{P}'_S = (\mathcal{O}', S, \mathcal{G})$, our version of GRAPHPLAN computes these layer memberships by using the following fixpoint computation. The layer memberships of all facts and actions are initialized to $\infty$. For each action, there is also a counter, which is initialized to 0. Then, fact layer 0 is built implicitly by setting the layer membership of all facts $f \in S$ to 0. Each time when a fact $f$ gets its layer membership set, all actions of which $f$ is a precondition get their counter incremented. As soon as the counter for an action $o$ reaches the total number of $o$'s preconditions, $o$ is put to a list of scheduled actions for the current layer. After a fact layer $i$ is finished, all actions scheduled for step $i$ have their layer membership set to $i$, and their adds, if not already present, are put to the list of scheduled facts for the next fact layer at time step $i + 1$. Having finished with action layer $i$, all scheduled facts at step $i + 1$ have their membership set, and so on. The process continues until all goals have a layer membership lower than $\infty$. It should be noticed here that this view of planning graph building corresponds closely to the computation of the weight values in HSP. Those can be computed by applying the actions in layers as above, updating weight values and propagating the changes each time an action comes in, and stopping when no changes occur in a layer. Having finished the relaxed version of planning graph building, a similarly trivial version of GRAPHPLAN's solution extraction mechanism is invoked. See Figure 2.

Instead of putting all goals into the top layer in GRAPHPLAN style, and then propagating them down by using NOOPs-first, each goal $g$ is simply put into a goal set $G_i$ located at $g$'s first layer $i$. Then, there is a for-next loop down from the top to the initial layer. At each layer $i$, an achieving action with layer membership $i - 1$ gets selected for each fact in the corresponding goal set. If there is more than one such achiever, a best one is picked according to the difficulty heuristic. The preconditions are put into their corresponding goal sets. Each time an action is selected, all of its adds are marked true at times $i$ and $i - 1$. The marker at time $i$ prevents achievers to be selected for facts that are already true

```
for i := 1, ..., m do
    G_i := {g ∈ G | layer-membership(g) = i}
endfor
for i := m, ..., 1 do
    for all g ∈ G_i, g not marked TRUE at time i do
        select an action o with g ∈ add(o) and layer membership i − 1, o's difficulty being minimal
        for all f ∈ pre(o), layer-membership(f) ≠ 0, f not marked TRUE at time i − 1 do
            G_{layer-membership(f)} := G_{layer-membership(f)} ∪ {f}
        endfor
        for all f ∈ add(o) do
            mark f as TRUE at times i − 1 and i
        endfor
    endfor
endfor
```

Figure 2: Relaxed plan extraction

anyway. Marking at time $i − 1$ assumes that actions are linearized in the order they get selected: A precondition that was achieved by an action ahead is not considered as a new goal.

## 5. A Novel Variation of Hill-climbing

In this section, we introduce FF's base search algorithm. We discuss the algorithm's theoretical properties regarding completeness, and derive FF's overall search strategy.

In the first HSP version (Bonet & Geffner, 1998), HSP1 as was used in the AIPS-1998 competition, the search strategy is a variation of hill-climbing, always selecting one best successor to the state it is currently facing. Because state evaluations are costly, we also chose to use local search, in the hope to reach goal states with as few evaluations as possible. We settled for a different search algorithm, an "enforced" form of hill-climbing, which combines local and systematic search. The strategy is motivated by the simple structure that the search spaces of our testing benchmarks tend to have.

### 5.1 Enforced Hill-climbing

Doing planning by heuristic forward search, the search space is the space of all reachable states, together with their heuristic evaluation. Now, evaluating states in our testing benchmarks with the heuristic defined by Equation 3, one often finds that the resulting search spaces are simple in structure, specifically, that local minima and plateaus tend to be small. For any search state, the next state with strictly better heuristic evaluation is usually only a few steps away (an example for this is the *Logistics* domain described in Section 8.1.1). Our idea is to perform exhaustive search for the better states. The algorithm is shown in Figure 3.

Like hill-climbing, the algorithm depicted in Figure 3 starts out in the initial state. Then, facing an intermediate search state $S$, a complete breadth first search starting out

```
initialize the current plan to the empty plan  <>
S := I
while h(S) ≠ 0 do
        perform breadth first search for a state S' with h(S') < h(S)
        if no such state can be found then
            output "Fail", stop
        endif
        add the actions on the path to S' at the end of the current plan
        S := S'
endwhile
```

Figure 3: The enforced hill-climbing algorithm.

from $S$ is invoked. This finds the closest better successor, i.e., the nearest state $S'$ with strictly better evaluation, or fails. In the latter case, the whole algorithm fails, in the former case, the path from $S$ to $S'$ is added to the current plan, and search is iterated. When a goal state—a state with evaluation 0—is reached, search stops.

Our implementation of breadth first search starting out from $S$ is standard, where states are kept in a queue. One search iteration removes the first state $S'$ from the queue, and evaluates it by running GRAPHPLAN. If the evaluation is better than that of $S$, search succeeds. Otherwise, the successors of $S'$ are put to the end of the queue. Repeated states are avoided by keeping a hash table of visited states in memory. If no new states can be reached anymore, breadth first search fails.

## 5.2 Completeness

If in one iteration breadth first search for a better state fails, then enforced hill-climbing stops without finding a solution. This can happen because once enforced hill-climbing has chosen to include an action in the plan, it never takes this decision back. The method is therefore only complete on tasks where no fatally wrong decisions can be made. These are the tasks that do not contain "dead ends."

**Definition 7 (Dead End)** Let $(\mathcal{O}, \mathcal{I}, \mathcal{G})$ be a planning task. A state $S$ is called a dead end iff it is reachable and no sequence of actions achieves the goal from it, i.e., iff $\exists P : S = Result(\mathcal{I}, P)$ and $\neg \exists P' : \mathcal{G} \subseteq Result(S, P')$.

Naturally, a task is called dead-end free if it does not contain any dead end states. We remark that being dead-end free implies solvability, as otherwise the initial state itself would already be a dead end.

**Proposition 4** Let $\mathcal{P} = (\mathcal{O}, \mathcal{I}, \mathcal{G})$ be a planning task. If $\mathcal{P}$ is dead-end free, then enforced hill-climbing will find a solution.

**Proof:** Assume enforced hill-climbing does not reach the goal. Then we have some intermediate state $S = Result(\mathcal{I}, P)$, $P$ being the current plan, where breadth first search can not improve on the situation. Now, $h(S) > 0$ as search has not stopped yet. If there was a path from $S$ to some goal state $S'$, then complete breadth first search would find that path, obtain $h(S') = 0 < h(S)$, and terminate positively. Such a path can therefore not exist, showing that $S$ is a dead end state in contradiction to the assumption. □

We remark that Proposition 4 holds only when $h$ is a function from states to natural numbers including 0, where $h(S) = 0$ iff $\mathcal{G} \subseteq S$. The proposition identifies a class of planning tasks where we can safely apply enforced hill-climbing. Unfortunately, it is PSPACE-hard to decide whether a given planning task belongs to that class.

**Definition 8** *Let DEADEND-FREE denote the following problem:*

Given a planning task $\mathcal{P} = (\mathcal{O}, \mathcal{I}, \mathcal{G})$, is $\mathcal{P}$ dead-end free?

**Theorem 3** *Deciding DEADEND-FREE is PSPACE-complete.*

**Proof:** Hardness is proven by polynomially reducing PLANSAT (Bylander, 1994)—the decision problem of whether $\mathcal{P}$ is solvable—to the problem of deciding DEADEND-FREE. We simply add an operator to $\mathcal{O}$ that is executable in all states, and re-establishes the initial state.

$$\mathcal{O}_1 := \mathcal{O} \cup \{o_I := \langle \emptyset, \mathcal{I}, \bigcup_{o \in \mathcal{O}} \text{add}(o) \setminus \mathcal{I} \rangle\}$$

Applying $o_I$ to any state reachable in $\mathcal{P}$ leads back to the initial state: all facts that can ever become true are removed, and those in the initial state are added. Now, the modified problem $\mathcal{P}_1 = (\mathcal{O}_1, \mathcal{I}, \mathcal{G})$ is dead-end free iff $\mathcal{P}$ is solvable. From left to right, if $\mathcal{P}_1$ is dead-end free, then it is solvable, which implies that $\mathcal{P}$ is solvable, as we have not added any new possibility of reaching the goal. From right to left, if $\mathcal{P}$ is solvable, then also is $\mathcal{P}_1$, by the same solution plan $P$. One can then, from all states in $\mathcal{P}_1$, achieve the goal by going back to the initial state with the new operator, and executing $P$ thereafter.

Membership in PSPACE follows from the fact that PLANSAT and its complement are both in PSPACE. A non-deterministic algorithm that decides the complement of DEADEND-FREE and that needs only polynomial space can be specified as follows. Guess a state $S$. Verify in polynomial space that $S$ is reachable from the initial state. Further, verify that the goal cannot be reached from $S$. If this algorithm succeeds, it follows that the instance is not dead-end free—since $S$ constitutes a dead end. This implies that DEADEND-FREE is in NPSPACE, and hence in PSPACE. □

Though we can not efficiently decide whether a given task is dead-end free, there are easily testable sufficient criteria in the literature. Johnsson et al. (2000) define a notion of *symmetric* planning tasks, which is sufficient for dead-end freeness, but co-NP-complete. They also give a polynomial sufficient criterion for symmetry. This is, however, very trivial. Hardly any of the current benchmarks fulfills it. Koehler and Hoffmann (2000a) have defined notions of *invertible* planning tasks—sufficient for dead-end freeness, and *inverse actions*—sufficient for invertibility, under certain restrictions. The existence of inverse actions, and sufficient criteria for the additional restrictions, can be decided in polynomial time. Many benchmark tasks do, in fact, fulfill those criteria and can thus efficiently be proven dead-end free.

One could adopt Koehler and Hoffmann's methodology, and use the existence of inverse actions to recognize dead-end free tasks. If the test fails, one could then employ a different search strategy than enforced hill-climbing. We have two reasons for *not* going this way:

- Even amongst our benchmarks, there are tasks that do not contain inverse actions, but are nevertheless dead-end free. An example is the *Tireworld* domain, where enforced hill-climbing leads to excellent results.

- Enforced hill-climbing can often quite successfully solve tasks that do contain dead ends, as it does not necessarily get caught in one. Examples for that are contained in the *Mystery* and *Mprime* domains, which we will look at in Section 8.2.1.

The observation that forms the basis for our way of dealing with completeness is the following. If enforced hill-climbing can not solve a planning task, it usually fails very quickly. One can then simply switch to a different search algorithm. We have experimented with randomizing enforced hill-climbing, and doing a restart when one attempt failed. This didn't lead to convincing results. Though we tried a large variety of randomization strategies, we did not find a planning task in our testing domains where one randomized restart did significantly better than the previous one, i.e., all attempts suffered from the same problems. The tasks that enforced hill-climbing does not solve right away are apparently so full of dead ends that one can not avoid those dead ends at random. We have therefore arranged our overall search strategy in FF as follows:

1. Do enforced hill-climbing until the goal is reached or the algorithm fails.

2. If enforced hill-climbing failed, skip everything done so far and try to solve the task by a complete heuristic search algorithm. In the current implementation, this is what Russel and Norvig (1995) term *greedy best-first* search. This strategy simply expands all search nodes by increasing order of goal distance estimation.

To summarize, FF uses enforced hill-climbing as the base search method, and a complete best-first algorithm to deal with those special cases where enforced hill-climbing has run into a dead end and failed.

## 6. Pruning Techniques

In this section, we introduce two heuristic techniques that can, in principle, be used to prune the search space in any forward state space search algorithm:

1. *Helpful actions* selects a set of promising successors to a search state. As we will demonstrate in Section 8.3, the heuristic is crucial for FF's performance on many domains.

2. *Added goal deletion* cuts out branches where some goal has apparently been achieved too early. Testing the heuristic, we found that it can yield savings on tasks that contain goal orderings, and has no effect on tasks that don't.

Both techniques are obtained as a side effect of using GRAPHPLAN as a heuristic estimator in the manner described in Section 4. Also, both of them do *not* preserve completeness of any hypothetical forward search. In the context of our search algorithm, we integrate them such that they prune the search space in the single enforced hill-climbing try—which is not complete in general anyway—and completely turn them off during best-first search, if enforced hill-climbing failed.

## 6.1 Helpful Actions

To a state $S$, we define a set $H(S)$ of actions that seem to be most promising among the actions applicable in $S$. The technique is derived by having a closer look at the relaxed plans that GRAPHPLAN extracts on search states in our testing tasks. Consider the *Gripper* domain, as it was used in the 1998 AIPS planning competition. There are two rooms, A and B, and a certain number of balls, which are all in room A initially and shall be moved into room B. The planner controls a robot, which changes rooms via the **move** operator, and which has two grippers to **pick** or **drop** balls. Each gripper can hold only one ball at a time. We look at a small task where 2 balls must be moved into room B. Say the robot has already **pick**ed up both balls, i.e., in the current search state, the robot is in room A, and each gripper holds one ball. There are three applicable actions in this state: **move** to room B, or **drop** one of the balls back into room A. The relaxed solution that our heuristic extracts is the following.

$<$ { **move** A B },
    { **drop** ball1 B left,
       **drop** ball2 B right } $>$

This is a parallel relaxed plan consisting of two time steps. The action set selected at the first time step contains the only action that makes sense in the state at hand, **move** to room B. We therefore pursue the idea of restricting the action choice in any planning state to only those actions that are selected in the first time step of the relaxed plan. We call these the actions that seem to be helpful. In the above example state, this strategy cuts down the branching factor from three to one.

Sometimes, restricting oneself to only the actions that are selected by the relaxed planner can be too much. Consider the following *Blocksworld* example. Say we use the well known representation with four operators, **stack**, **unstack**, **pickup** and **putdown**. The planner controls a single robot arm, and the operators can be used to **stack** one block on top of another one, **unstack** a block from another one, **pickup** a block from the table, or **put** a block that the arm is holding **down** onto the table. Initially, the arm is holding block C, and blocks A and B are on the table. The goal is to stack A onto B. Started on this state, relaxed GRAPHPLAN will return one out of the following three time step optimal solutions.

$<$ { **putdown** C },
    { **pickup** A },
    { **stack** A B } $>$

or

< { **stack** C A },
  { **pickup** A },
  { **stack** A B } >

or

< { **stack** C B },
  { **pickup** A },
  { **stack** A B } >

All of these are valid relaxed solutions, as in the relaxation it does not matter that **stack**ing C onto A or B deletes facts that we still need. If C is on A, we can not **pickup** A anymore, and if C is on B, we can not **stack** A onto B anymore.

The first action in each relaxed plan is only inserted to get rid of C, i.e., free the robot arm, and from the point of view of the relaxed planner, all of the three starting actions do the job. Thus the relaxed solution extracted might be any of the three above. If it happens to be the second or third one, then we lose the path to an optimal solution by restricting ourselves to the corresponding actions, **stack** C A or **stack** C B. Therefore, we define the set $H(S)$ of helpful actions to a state $S$ as follows.

$$H(S) := \{o \mid \mathrm{pre}(o) \subseteq S, add(o) \cap G_1(S) \neq \emptyset\} \tag{5}$$

Here, $G_1(S)$ denotes the set of goals that is constructed by relaxed GRAPHPLAN at time step 1—one level ahead of the initial layer—when started on the task $(\mathcal{O}', S, \mathcal{G})$. In words, we consider as helpful actions all those applicable ones, which add at least one goal at the first time step. In the above *Blocksworld* example, freeing the robot arm is among these goals, which causes all the three starting actions to be helpful in the initial state, i.e., to be elements of $H(\mathcal{I})$. In the above *Gripper* example, the modification does not change anything.

The notion of helpful actions shares some similarities with what Drew McDermott calls the *favored actions* (McDermott, 1996, 1999), in the context of computing *greedy regression graphs* for heuristic estimation. In a nutshell, greedy regression graphs backchain from the goals until facts are reached that are contained in the current state. Amongst other things, the graphs provide an estimation of which actions might be useful in getting closer to the goal: Those applicable ones which are members of the *effective subgraph*, which is the minimal cost subgraph achieving the goals.

There is also a similarity between the helpful actions heuristic and what is known as *relevance* from the literature (Nebel, Dimopoulos, & Koehler, 1997). Consider a *Blocksworld* task where hundreds of blocks are on the table initially, but the goal is only to stack one block A on top of another block B. The set $H(\mathcal{I})$ will in this case contain only the single action **pickup** A, throwing away all those applicable actions moving around blocks that are not mentioned in the goal, i.e., throwing away all those actions that are irrelevant. The main difference between the helpful actions heuristic and the concept of relevance is that relevance in the usual sense refers to what is useful for solving the whole task. Being helpful, on the other hand, refers to something that is useful *in the next step*. This has the disadvantage that the helpful things need to be recomputed for each search state, but the advantage that possibly far less things are helpful than are relevant. In our specific setting, we get the helpful actions for free anyway, as a side effect of running relaxed GRAPHPLAN.

We conclude this subsection with an example showing that helpful actions pruning does *not* preserve completeness, and a few remarks on the current integration of the technique into our search algorithm.

### 6.1.1 COMPLETENESS

In the following short example, the helpful actions heuristic prunes out all solutions from the state space. Say the initial state is $\{B\}$, the goals are $\{A, B\}$, and there are the following actions:

| name | | (pre, | add, | del) |
|------|---|-------|------|------|
| $\mathbf{op}A_1$ | $=$ | $(\emptyset,$ | $\{A\},$ | $\{B\})$ |
| $\mathbf{op}A_2$ | $=$ | $(\{P_A\},$ | $\{A\},$ | $\emptyset)$ |
| $\mathbf{op}P_A$ | $=$ | $(\emptyset,$ | $\{P_A\},$ | $\emptyset)$ |
| $\mathbf{op}B_1$ | $=$ | $(\emptyset,$ | $\{B\},$ | $\{A\})$ |
| $\mathbf{op}B_2$ | $=$ | $(\{P_B\},$ | $\{B\},$ | $\emptyset)$ |
| $\mathbf{op}P_B$ | $=$ | $(\emptyset,$ | $\{P_B\},$ | $\emptyset)$ |

In this planning task, there are two ways of achieving the missing goal $A$. One of these, $\mathbf{op}A_1$, deletes the other goal $B$. The other one, $\mathbf{op}A_2$, needs the precondition $P_A$ to be achieved first by $\mathbf{op}P_A$, and thus involves using *two* planning actions instead of *one* in the first case. Relaxed GRAPHPLAN recognizes only the first alternative, as it's the only time step optimal one. The set of goals at the single time step created by graph construction is

$$G_1(\mathcal{I}) = \{ A, B \}$$

This gives us two helpful actions, namely

$$H(\mathcal{I}) = \{ \mathbf{op}A_1, \mathbf{op}B_1 \}$$

One of these, $\mathbf{op}B_1$, does not cause any state transition in the initial state. The other one, $\mathbf{op}A_1$, leads to the state where only $A$ is true. To this state, we obtain the same set of helpful actions, containing, again, $\mathbf{op}A_1$ and $\mathbf{op}B_1$. This time, the first action causes no state transition, while the second one leads us back to the initial state. Helpful actions thus cuts out the solutions from the state space of this example task. We remark that the task is dead-end free—one can always reach $A$ and $B$ by applying $\mathbf{op}P_A$, $\mathbf{op}A_2$, $\mathbf{op}P_B$, and $\mathbf{op}B_2$—and that one can easily make the task invertible without changing the behavior.

In STRIPS domains, one could theoretically overcome the incompleteness of helpful actions pruning by considering not only the first relaxed plan that GRAPHPLAN finds, but computing a kind of union over all relaxed plans that GRAPHPLAN could possibly find, when allowing non time step optimal plans. More precisely, in a search state $S$, consider the relaxed task $(\mathcal{O}', S, \mathcal{G})$. Extend the relaxed planning graph until fact level $|\mathcal{O}'|$ is reached. Set a goal set $G_{|\mathcal{O}'|}$ at the top fact level to $G_{|\mathcal{O}'|} := \mathcal{G}$. Then, proceed from fact level $|\mathcal{O}'| - 1$ down to fact level 1, where, at each level $i$, a set $G_i$ of goals is generated as the union of $G_{i+1}$ with the preconditions of all actions in level $i$ that add at least one fact in $G_{i+1}$. Upon termination, define as helpful all actions that add at least one fact in $G_1$. It can be proven that, this way, the starting actions of all optimal solutions from $S$ are considered helpful. However, in all our STRIPS testing domains, the complete method always selects *all* applicable actions as helpful.

### 6.1.2 Integration into Search

As has already been noted at the very beginning of this section, we integrate helpful actions pruning into our search algorithm by only applying it during the single enforced hill-climbing try, leaving the complete best-first search algorithm unchanged (see Section 5). Facing a state $S$ during breadth first search for a better state in enforced hill-climbing, we look only at those successors generated by $H(S)$. This renders our implementation of enforced hill-climbing incomplete even on invertible planning tasks. However, in all our testing domains, the tasks that cannot be solved by enforced hill-climbing using helpful actions pruning are exactly those that cannot be solved by enforced hill-climbing anyway.

## 6.2 Added Goal Deletion

The second pruning technique that we introduce in this section is motivated by the observation that in some planning domains there are goal ordering constraints, as has been recognized by quite a number of researchers in the past (Irani & Cheng, 1987; Drummond & Currie, 1989; Joslin & Roach, 1990). In our experiments on tasks with goal ordering constraints, FF's base architecture sometimes wasted a lot of time achieving goals that needed to be cared for later on. We therefore developed a heuristic to inform search about goal orderings.

The classical example for a planning domain with goal ordering constraints is the well known *Blocksworld*. Say we have three blocks A, B and C on the table initially, and want to stack them such that we have B on top of C, and A on top of B. Obviously, there is not much point in stacking A on B first. Now, imagine a forward searching planner confronted with a search state $S$, where some goal $G$ has just been achieved, i.e., $S$ resulted from some other state by applying an action $o$ with $G \in \text{add}(o)$. What one can ask in a situation like this is, was it a good idea to achieve $G$ right now? Or should some other goal be achieved first? Our answer is inspired by recent work of Koehler and Hoffmann (2000a), which argues that achieving $G$ should be postponed if the remaining goals can not be achieved without destroying $G$ again. Of course, finding out about this involves solving the remaining planning task. However, we can arrive at a very simple but—in our testing domains—surprisingly accurate approximation by using the relaxed plan that GRAPHPLAN generates for the state $S$. The method we are using is as simple as this: If the relaxed solution plan, $P$, that GRAPHPLAN generates for $S$, contains an action $o$, $o \in P$, that deletes $G$ ($G \in \text{del}(o)$ in $o$'s non-relaxed version), then we remove $S$ from the search space, i.e., do not generate any successors to $S$. We call this method the *added goal deletion* heuristic.

Let us exemplify the heuristic with the above *Blocksworld* example. Say the planner has just achieved $on(A,B)$, but with $on(B,C)$ still being false, i.e., we are in the situation where A is on top of B, and B and C are standing on the table. The relaxed solution that GRAPHPLAN finds to this situation is the following.

< { **unstack** A B },
  { **pickup** B },
  { **stack** B C } >

The goal *on*(A,B), which has just been achieved, gets deleted by the first action **unstack** A B. Consequently, we realize that **stack**ing A onto B right now was probably a bad idea, and prune this possibility from the search space, which results in a solution plan that **stack**s B onto C first.

Like in the preceding subsection, we conclude with an example showing that pruning search states in the manner described above does not preserve completeness, and with a few remarks on our current search algorithm implementation.

### 6.2.1 COMPLETENESS

In the following small example, one of the goals *must* be destroyed temporarily in order to achieve the other goal. This renders the planning task unsolvable when one is using the added goal deletion heuristic. Say the initial state is empty, the goals are $\{A, B\}$, and there are the following actions:

| name | | (pre, | add, | del) |
|------|---|-------|------|------|
| **op**$A$ | = | $(\emptyset,$ | $\{A\},$ | $\emptyset)$ |
| **op**$B$ | = | $(\{A\},$ | $\{B\},$ | $\{A\})$ |

All solutions to this task need to apply **op**$A$, use **op**$B$ thereafter, and re-establish $A$. The crucial point here is that $A$ *must* be temporarily destroyed. The added goal deletion heuristic is not adequate for such planning tasks. The example is dead-end free, and one can easily make the scenario invertible without changing the behavior of the heuristic.

Unlike for helpful actions, completeness can not be regained by somehow enumerating all relaxed plans to a situation. In the above example, when $A$ has been achieved but $B$ is still FALSE, then all relaxed plans contain **op**$B$, deleting $A$.

### 6.2.2 INTEGRATION INTO SEARCH

We use the added goal deletion heuristic in a way similar to the integration of the helpful actions heuristic. As indicated at the very beginning of the section, it is integrated into the single enforced hill-climbing try that search does, and completely turned off during best-first search, in case enforced hill-climbing didn't make it to the goal.

We also use another goal ordering technique, taken from the literature. One of the most common approaches to dealing with goal orderings is trying to recognize them in a preprocessing phase, and then use them to prune fractions of the search space during planning (Irani & Cheng, 1987; Cheng & Irani, 1989; Joslin & Roach, 1990). This is also the basic principle underlying the so-called "goal agenda" approach (Koehler, 1998). For our system, we have implemented a slightly simplified version of the goal agenda algorithm, and use it to further enhance performance. A very short summary of what happens is this.

In a preprocessing phase, the planner looks at all pairs of goals and decides heuristically whether there is an ordering constraint between them. Afterwards, the goal set is split into a totally ordered series of subsets respecting these orderings. These are then fed to enforced hill-climbing in an incremental manner. Precisely, if $G_1, \ldots, G_n$ is the ordered series of subsets, enforced hill-climbing gets first started on the original initial state and $G_1$. If that works out, search ends in some state $S$ satisfying the goals in $G_1$. Enforced hill-climbing is then called again on the new starting state $S$ and the larger goal set $G_1 \cup G_2$. From a state satisfying this, search gets started for the goals $G_1 \cup G_2 \cup G_3$, and so on. The incremental, or *agenda-driven*, planning process can be applied to any planner, in principle, and preserves completeness only on dead-end free tasks (Koehler & Hoffmann, 2000a), i.e., again, we have an enhancement that loses completeness in general. Thus, we use the goal agenda only in enforced hill-climbing, leaving the complete best-first search phase unchanged.

The goal agenda technique yields runtime savings in domains where there are ordering constraints between the goals. In our testing suite, these are the *Blocksworld* and the *Tireworld*. In planning tasks without ordering constraints, the series of subsets collapses into a single entry, such that the agenda mechanism does not change anything there. The runtime taken for the pre-process itself was neglectible in all our experiments.

## 7. Extension to ADL

So far, we have restricted ourselves to planning tasks specified in the simple STRIPS language. We will now show how our approach can be extended to deal with ADL (Pednault, 1989) tasks, more precisely, with the ADL subset of PDDL (McDermott et al., 1998) that was used in the 2nd international planning systems competition (Bacchus, 2000). This involves dealing with arbitrary function-symbol free first order logic formulae, and with conditional effects. Our extension work is divided into the following four subareas:

1. Apply a preprocessing approach to the ADL domain and task description, compiling the specified task down into a propositional normal form.

2. Extend the heuristic evaluation of planning states to deal with these normal form constructs.

3. Adjust the pruning techniques.

4. Adjust the search mechanisms.

### 7.1 Preprocessing an ADL Planning Task

FF's preprocessing phase is almost identical to the methodology that has been developed for the IPP planning system. For details, we refer the reader to the work that's been done there (Koehler & Hoffmann, 2000b), and give only the basic principles here.

The planner starts with a planning task specification given in the subset of PDDL defined for the AIPS-2000 planning competition (Bacchus, 2000). The input is a set of operator schemata, the initial state, and a goal formula. The initial state is simply a set of ground atoms, and the goal formula is an arbitrary first order logical formula using the relational symbols defined for the planning task. Any operator schema $o$ is defined by a list of parameters, a precondition, and a list of effects. Instantiating the parameters yields, just like STRIPS tasks are usually specified, the actions to the schema. The precondition is an arbitrary (first order) formula. For an action to be applicable in a given state $S$, its instantiation of this formula must be satisfied in $S$. Each effect $i$ in the list has the form

$$\forall y_0^i, \ldots, y_{n^i}^i : (\psi^i(o), \mathrm{add}^i(o), \mathrm{del}^i(o))$$

Here, $y_0^i, \ldots, y_{n^i}^i$ are the effect parameters, $\psi^i(o)$ is the effect condition—again, an arbitrary formula—and $\mathrm{add}^i(o)$ and $\mathrm{del}^i(o)$ are the atomic add and delete effects, respectively. The atomic effects are sets of uninstantiated atoms, i.e., relational symbols containing variables. The semantics are that, if an instantiated action is executed, then, for each single effect $i$ in the list, and for each instantiation of its parameters, the condition $\psi^i(o)$ is evaluated. If $\psi^i(o)$ holds in the current state, then the corresponding instantiations of the atoms in $\mathrm{add}^i(o)$ are added to the state, and the instantiations of atoms in $\mathrm{del}^i(o)$ are removed from the state.

In FF's heuristic method, each single state evaluation can involve thousands of operator applications—building the relaxed planning graph, one needs to determine all applicable actions at each single fact layer. We therefore invest the effort to compile the operator descriptions down into a much simpler propositional normal form, such that heuristic evaluation can be implemented efficiently. Our final normal form actions $o$ have the following format.

Precondition: $\mathrm{pre}(o)$
Effects:      $(\mathrm{pre}^0(o), \mathrm{add}^0(o), \mathrm{del}^0(o)) \ \wedge$
              $(\mathrm{pre}^1(o), \mathrm{add}^1(o), \mathrm{del}^1(o)) \ \wedge$
              .
              .
              .
              $(\mathrm{pre}^m(o), \mathrm{add}^m(o), \mathrm{del}^m(o))$

The precondition is a set of ground atoms. Likewise, the effect conditions $\mathrm{pre}^i(o)$ of the single effects are restricted to be ground atoms. We also represent the goal state as a set of atoms. Thus, we compile away everything except the conditional effects. Compiling away the logical formulae involves transforming them into DNF, which causes an exponential blow up in general. In our testing domains, however, we found that this transformation can be done in reasonable time. Concerning the conditional effects, those can not be compiled away without another exponential blow up, given that we want to preserve solution length. This was proven by Nebel (2000). As we will see, conditional effects can efficiently be integrated into our algorithmic framework, so there is no need for compiling them away. The compilation process proceeds as follows:

1. Determine predicates that are *static*, in the sense that no operator has an effect on them. Such predicates are a common phenomenon in benchmark tasks. An example are the (in-city ?l ?c) facts in *Logistics* tasks: Any location ?l stays, of course, located within the same city ?c throughout the whole planning process. We recognize static predicates by a simple sweep over all operator schemata.

2. Transform all formulae into quantifier-free DNF. This is subdivided into three steps:

   (a) Pre-normalize all logical formulae. Following Gazen and Knoblock (1997), this process expands all quantifiers, and translates negations. We end up with formulae that are made up out of conjunctions, disjunctions, and atoms containing variables.

   (b) Instantiate all parameters. This is simply done by instantiating all operator and effect parameters with all type consistent constants one after the other. The process makes use of knowledge about static predicates, in the sense that the instantiated formulae can often be simplified (Koehler & Hoffmann, 2000b). For example, if an instantiated static predicate $(p\ \vec{a})$ occurs in a formula, and that instantiation is *not* contained in the initial state, then $(p\ \vec{a})$ can be replaced with FALSE.

   (c) Transform formulae into DNF. This is postponed until after instantiation, because it can be costly, so it should be applied to as small formulae as possible. In a fully instantiated formula, it is likely that many static or one-way predicate occurrences can be replaced by TRUE or FALSE, resulting in a much simpler formula structure.

3. Finally, if the DNF of any formula contains more than one disjunct, then the corresponding effect, operator, or goal condition gets split up in the manner proposed by Gazen and Knoblock (1997).

## 7.2 Relaxed GRAPHPLAN with Conditional Effects

We now show how our specialized GRAPHPLAN implementation, as was described in Section 4.3, is changed to deal with ADL constructs. Building on our normalized task representation, it suffices to take care of conditional effects.

### 7.2.1 Relaxed Planning Graphs with Conditional Effects

Our encoding of planning graph building for relaxed tasks almost immediately carries over to ADL actions in the above propositional normal form. One simply needs to keep an additional layer membership value for all *effects* of an action. The layer membership of an effect indicates the first layer where all its effect conditions plus the corresponding action's preconditions are present. To compute these membership integers in an efficient manner, we keep a counter for each effect $i$ of an action $o$, which gets incremented each time a condition $c \in \text{pre}^i(o)$ becomes present, *and* each time a precondition $p \in \text{pre}(o)$ of the corresponding action becomes present. The effect gets its layer membership set as soon as its counter reaches $|\text{pre}^i(o)| + |\text{pre}(o)|$. The effect's add effects $\text{add}^i(o)$ are then scheduled for the next layer. The process is iterated until all goals are reached the first time.

### 7.2.2 Relaxed Plan Extraction with Conditional Effects

The relaxed plan extraction mechanism for ADL differs from its STRIPS counterpart in merely two little details. Instead of selecting achieving actions, the extraction mechanism selects achieving effects. Once an effect $i$ of action $o$ is selected, all of its effect conditions plus $o$'s preconditions need to be put into their corresponding goal sets. Afterwards, not only the effect's own add effects $\text{add}^i(o)$ are marked TRUE at the time being, but also the added facts of all effects that are *implied*, i.e., those effects $j$ of $o$ with $\text{pre}^j(o) \subseteq \text{pre}^i(o)$ (in particular, this will be the unconditional effects of $o$, which have an empty effect condition).

## 7.3 ADL Pruning Techniques

Both pruning techniques from Section 6 easily carry over to actions with conditional effects.

### 7.3.1 HELPFUL ACTIONS

For STRIPS, we defined as helpful all applicable actions achieving at least one goal at time step 1, cf. Section 6.1. For our ADL normal form, we simply change this to *all applicable actions having an appearing effect that achieves a goal at time step 1*, where an effect *appears* iff its effect condition is satisfied in the current state.

$$H(S) := \{o \mid \text{pre}(o) \subseteq S, \exists i : \text{pre}^i(o) \subseteq S \wedge \text{add}^i(o) \cap G_1(S) \neq \emptyset\} \qquad (6)$$

### 7.3.2 ADDED GOAL DELETION

Originally, we cut off a state $S$ if one of the *actions* selected for the relaxed plan to $S$ deleted a goal $A$ that had just been achieved, cf. Section 6.2. We now simply take as criterion the *effects* that are selected for the relaxed plan, i.e., a state is cut off if one of the effects selected for its relaxed solution deletes a goal $A$ that has just been achieved.

## 7.4 ADL State Transitions

Finally, for enabling the search algorithms to handle our propositional ADL normal form, it is sufficient to redefine the *state transition function*. Forward search, no matter if it does hill-climbing, best-first search, or whatsoever, always faces a completely specified search state.[3] It can therefore compute exactly the effects of executing a context dependent action. Following Koehler et al.(1997), we define our ADL state transition function $Res$, mapping states and ADL normal form actions to states, as follows.

$$Res(S,o) = \left\{ \begin{array}{ll} (S \cup A(S,o)) \setminus D(S,o) & \text{if } \text{pre}(o) \subseteq S \\ \text{undefined} & \text{otherwise} \end{array} \right.$$

with

$$A(S,o) = \bigcup_{\text{pre}^i(o) \subseteq S} \text{add}^i(o) \qquad \text{and} \qquad D(S,o) = \bigcup_{\text{pre}^i(o) \subseteq S} \text{del}^i(o) \ .$$

---

3. This holds if the initial state is completely specified and all actions are deterministic, which we both assume.

## 8. Performance Evaluation

We have implemented the methodology presented in the preceding sections in C.[4] In this section, we evaluate the performance of the resulting planning system. Empirical data is divided into three subareas:

1. The FF system took part in the fully automated track of the 2nd international planning systems competition, carried out alongside with AIPS-2000 in Breckenridge, Colorado. We review the results, demonstrating FF's good runtime and solution length behavior in the competition. We also give some intuitions on why FF behaves the way it does.

2. From our own experiments, we present some of the results that we have obtained in domains that were not used in the AIPS-2000 competition. First, we briefly summarize our findings in some more domains where FF works well. Then, to illustrate our intuitions on the reasons for FF's performance, we give a few examples of domains where the approach is less appropriate.

3. We finally present a detailed comparison of FF's performance to that of HSP, in the sense that we investigate which differences between FF and HSP lead to which performance results.

### 8.1 The AIPS-2000 Planning Systems Competition

From March to April 2000, the 2nd international planning systems competition, organized by Fahiem Bacchus, was carried out in the general setting of the AIPS-2000 conference in Breckenridge, Colorado. There were two main tracks, one for fully-automated planners, and one for hand-tailored planners. Both tracks were divided into five parts, each one concerned with a different planning domain. Our FF system took part in the fully automated track. In the competition, FF demonstrated runtime behavior superior to that of the other fully automatic planners and was therefore granted "Group A distinguished performance Planning System" (Bacchus & Nau, 2001). It also won the Schindler Award for the first place in the Miconic 10 Elevator domain, ADL track. In this section, we briefly present the data collected in the fully automated track, and give, for each domain, some intuitions on the reasons for FF's behavior. The reader should be aware that the competition made no distinction between optimal and suboptimal planners, putting together the runtime curves for both groups. In the text to each domain, we state which planners found optimal solutions, and which didn't. Per planning task, all planners were given half an hour running time on a 500 MHz Pentium III with 1GB main memory. If no solution was found within these resource bounds, the planner was declared to have failed on the respective task.

### 8.1.1 THE *Logistics* DOMAIN

The first two domains that were used in the competition were the *Logistics* and *Blocksworld* domains. We first look at the former. This is a classical domain, involving the transportation of packets via trucks and airplanes. Figure 4 shows the runtime curves of those planners that were able to scale to bigger instances in the competition.

---

4. The source code is available in an online appendix, and can be downloaded from the FF Homepage at http://www.informatik.uni-freiburg.de/~ hoffmann/ff.html.

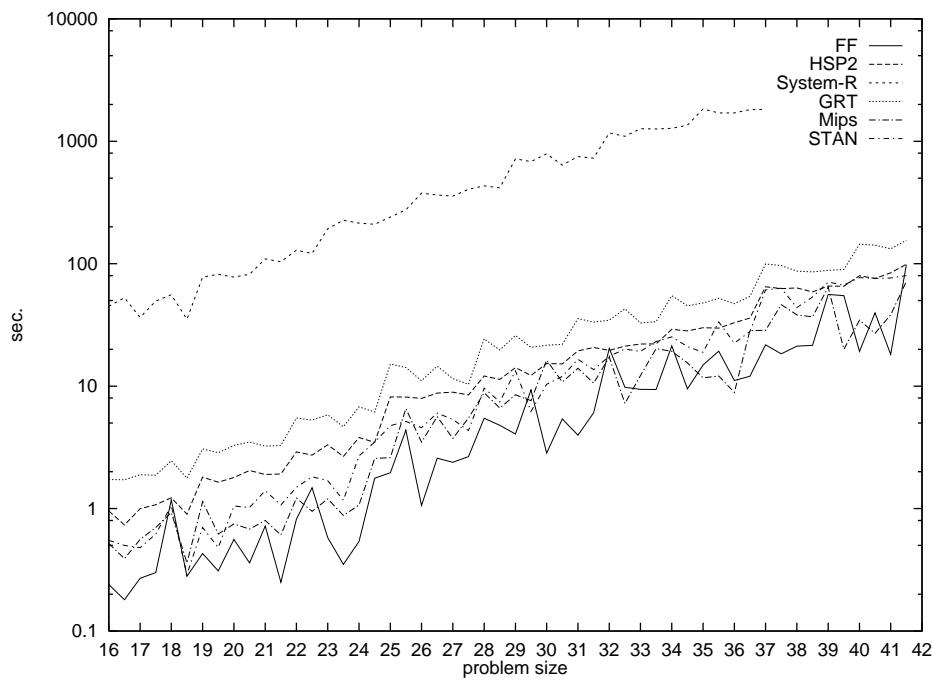

Figure 4: Runtime curves on large *Logistics* instances for those six planners that could scale up to them. Time is shown on a logarithmic scale.

The *Logistics* tasks were subdivided into two sets of instances, the easy and the harder ones. Those planners that did well on all of the easy instances were also run on the harder set. These planners were FF, HSP2 (Bonet & Geffner, 1998, 1999), System-R, GRT (Refanidis & Vlahavas, 1999), Mips (Edelkamp, 2000), and STAN (Long & Fox, 1999; Fox & Long, 2001). Two observations can be made:

1. System-R does significantly worse than the other planners.

2. The better planners all behave quite similar, with FF and Mips tending to be the fastest.

Note also that times are shown on a logarithmic scale, so we are *not* looking at linear time *Logistics* planners. Concerning solution plan length, we do not show a figure here. None of the shown planners guarantees the returned plans to be optimal. It turns out that STAN finds the shortest plans on most instances. System-R finds significantly longer plans than the others, ranging from 178% to 261% of STAN's plan lengths, with an average of 224%. The lengths of FF's plans are within 97% to 115% of STAN's plan lengths, with an average length of 105%. Concerning FF's good runtime behavior, we think that there are mainly two reasons for that:

1. In all iterations of enforced hill-climbing, breadth first search finds a state with better evaluation at very small depths (motivating our search algorithm, cf. Section 5.1). In most cases, the next better successor is at depth 1, i.e., a direct one. There are some cases where the shallowest better successor is at depth 2, and only very rarely breadth first needs to go down to depth 3. These observations are independent of task size.

2. The helpful actions heuristic prunes large fractions of the search space. Looking at the states that FF encounters during search, only between 40 and 5 percent of all of a state's successors were considered helpful in our experiments, with the tendency that the larger the task, the less helpful successors there are.

There is a theoretical note to be made on the first observation. With the common representation of *Logistics* tasks, the following can be proven. Let $d$ be the maximal distance between two locations, i.e., the number of move actions a mobile needs to take to get from one location to another. Using a heuristic function that assigns to each state the length of an *optimal* relaxed solution as the heuristic value, the distance of each state to the next better evaluated state is maximal $d+1$. Thus, an algorithm that used enforced hill-climbing with an oracle function returning the length of an optimal relaxed solution would be polynomial on standard *Logistics* representations, given an upper limit to $d$. In the benchmarks available, mobiles can reach any location accessible to them in just one step, i.e., the maximal distance in those tasks is constantly $d = 1$. Also, FF's heuristic usually *does* find optimal, or close to optimal, relaxed solutions there, such that enforced hill-climbing almost never needs to look more than $d + 1 = 2$ steps ahead.

## 8.1.2 The *Blocksworld* Domain

The *Blocksworld* is one of the best known benchmark planning domains, where the planner needs to rearrange a bunch of blocks into a specified goal position, using a robot arm. Just like the *Logistics* tasks, the competition instances were divided into a set of easier, and of harder ones. Figure 5 shows the runtime curves of the planners that scaled to the harder ones.

System-R scales most steadily to the *Blocksworld* tasks used in the competition. In particular, it is the only planner that can solve *all* of those tasks. HSP2 solves some of the smaller instances, and FF solves about two thirds of the set. If FF succeeds on an instance, then it does so quite fast. For example, FF solves one of the size-50 tasks in 1.27 seconds, where System-R needs 892.31 seconds. None of the three planners finds optimal plans. On the tasks that HSP2 manages to solve, its plans are within 97% to 177% of System-R's plan lengths, with an average of 153%. On the tasks that FF manages to solve, its plans are within 83% to 108% of System-R's plan lengths, average 96%.

By experimenting with different configurations of FF, we found that the behavior of FF on these tasks is largely due to the goal ordering heuristics from Section 6.2. Goal distance estimates are not so good—the planner grabs a whole bunch of blocks with its single arm—and neither is the helpful actions heuristic—when the arm holds a block, all positions where the arm could possibly put the block are usually considered helpful. The goal agenda (Section 6.2.2), on the other hand, divides the tasks into small subtasks, and added goal deletion (Section 6.2) prevents the planner from putting blocks onto stacks where some block beneath still needs to be moved. However, in some cases achieving the goals from earlier entries in the goal agenda cuts off goals that are still ahead. Not aware of the blocks that it will need to stack for achieving goals ahead, the planner might put the current blocks onto stacks that need to be disassembled later on. If that happens with too many blocks—which depends more or less randomly on the specific task and the actions that the planner chooses—then the planner can not find its way out of the situation again. These are probably the instances that FF couldn't solve in the competition.

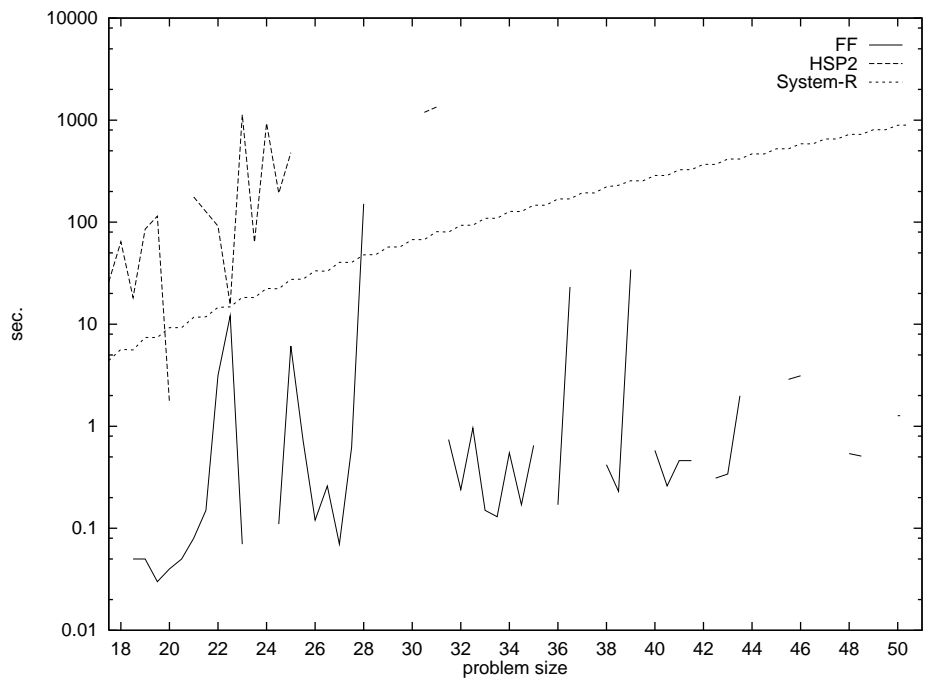

Figure 5: Runtime curves on large *Blocksworld* instances for those three planners that could scale up to them: FF, HSP2, and System-R. Time is shown an a logarithmic scale.

### 8.1.3 THE *Schedule* DOMAIN

In the *Schedule* domain, the planner is facing a bunch of objects to be worked on with a set of machines, i.e., the planner is required to create a job schedule in which the objects shall be assigned to the machines. The competition representation makes use of a simple form of quantified conditional effects. For example, if an object gets painted red, then that is its new color, and for all colors that it is currently painted in, it is not of that color anymore. Only a subset of the planners in the competition could handle this kind of conditional effects. Their runtime curves are shown in Figure 6.

Apart from those planners already seen, we have runtime curves in Figure 6 for IPP (Koehler et al., 1997), PropPlan, and BDDPlan (Hölldobler & Störr, 2000). FF outperforms the other planners by many orders of magnitude—remember that time is shown on a logarithmic scale. Concerning solution length, FF's plans tend to be slightly longer than the plans returned by the other planners on the smaller instances. Optimal plans are found by Mips, PropPlan, and BDDPlan. FF's plan lengths are within 175% of the optimal lengths, with an average of 116%. Only HSP sometimes finds longer plans than FF, being in a range from 62% to 117% of FF's plan lengths, 94% on average.

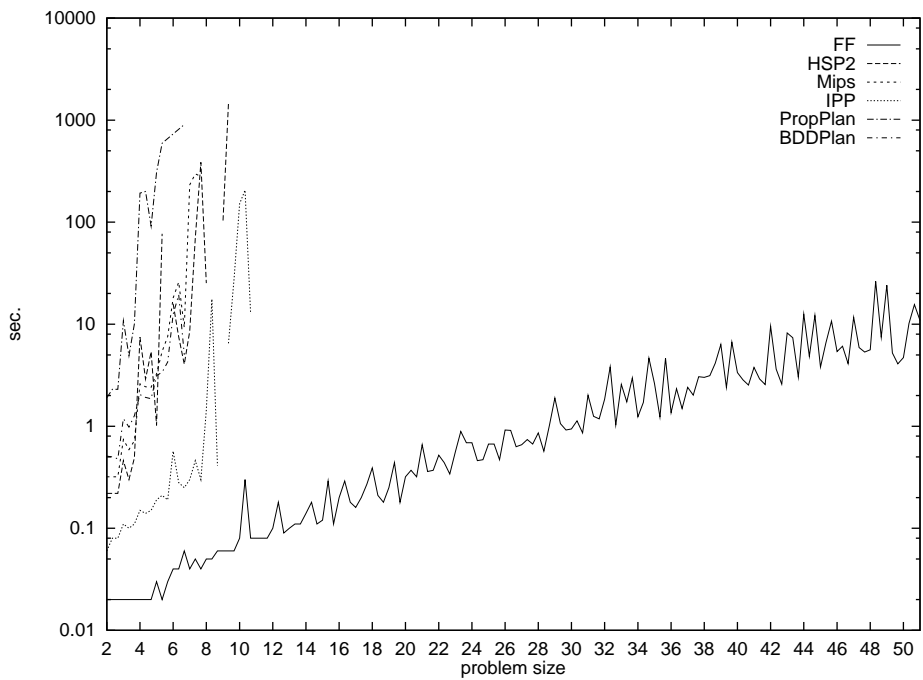

Figure 6: Runtime curves on *Schedule* instances for those planners that could handle conditional effects. Time is shown on a logarithmic scale.

Responsible for the outstanding runtime behavior of FF on the *Schedule* domain is, apparently, the helpful actions heuristic. Measuring, for some example states, the percentage of successors that were considered helpful, we usually found it was close to 2 percent, i.e., only two out of a hundred applicable actions were considered by the planner. For example, all of the 637 states that FF looks at for solving one of the size-50 tasks have 523030 successors altogether, where the sum of all *helpful* successors is only 7663. Also, the better successors, similar to the *Logistics* domain, lie at shallow depths. Breadth first search never goes deeper than three steps on the *Schedule* tasks in the competition suite. Finally, in a few experiments we ran for testing that, the goal agenda helped by about a factor 2 in terms of running time.

### 8.1.4 THE *Freecell* DOMAIN

The *Freecell* domain formalizes a solitaire card game that comes with Microsoft Windows. The largest tasks entered in the competition (size 13 in Figure 7) correspond directly to some real-world sized tasks, while in the smaller tasks, there are less cards to be considered. Figure 7 shows the runtime curves of the four best performing planners.

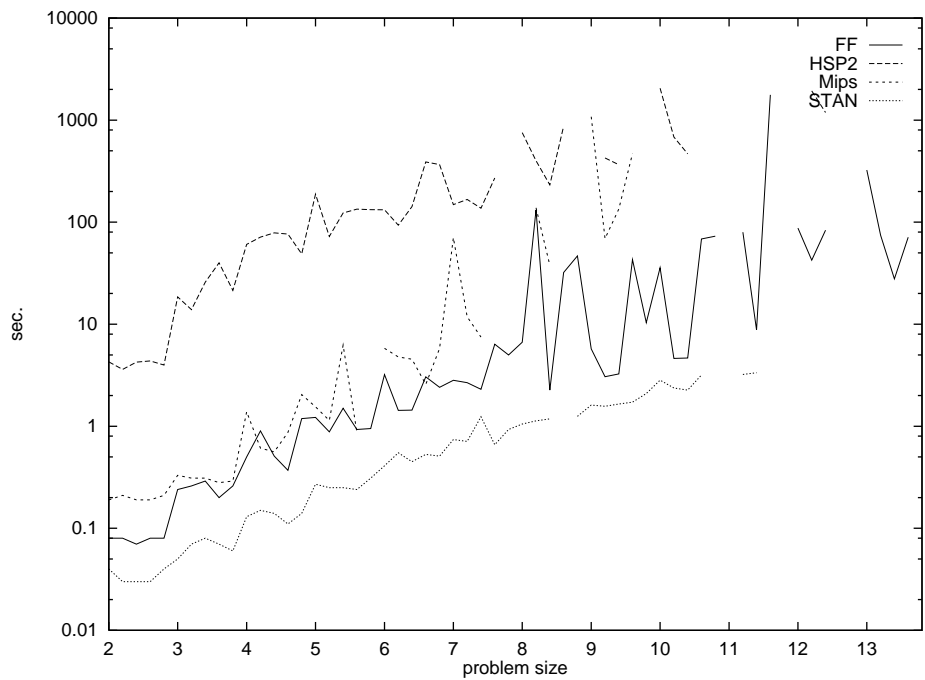

Figure 7: Runtime curves on *Freecell* tasks for those planners that scaled to bigger instances. Time is shown on a logarithmic scale.

From the group of the four best-scaling planners shown in Figure 7, HSP2 is the slowest, while STAN is the fastest planner. FF is generally second place, and has a lot of variation in its running times. On the other hand, FF is the only planner that is capable of solving the real-world tasks, size 13. It solves four out of five such tasks. None of the shown planners guarantees the found plans to be optimal, and none of the shown planners demonstrates superior performance concerning solution length. STAN produces unnecessarily long plans in a few cases. Precisely, on the tasks that both HSP and FF manage to solve, HSP's plan lengths are within a range of 74% to 126% of FF's plan lengths, average 95%. On tasks solved by both Mips and FF, plan lengths of Mips are within 69% to 128% of FF's lengths, average 101%. For STAN, the range is 65% to 318%, with 112% on average.

Concerning FF's runtime behavior, the big variation in running time as well as its capability of solving larger tasks both seem to result from the way the overall search algorithm is arranged. We observed the following. Those tasks that get solved by enforced hill-climbing are those that are solved fast. Sometimes, however, especially on the larger tasks, enforced hill-climbing runs into a dead end situation (no cards can be moved anymore). Then, the planner starts from scratch with complete best-first search, which takes more time, but can solve big instances quite reliably, as can be seen on the tasks of size 13. Helpful actions works moderately well, selecting around 70% of the available actions, and the better successors are usually close, but sometimes lie at depths of more than 5 steps.

### 8.1.5 THE *Miconic* DOMAIN

The final domain used in the competition comes from a real-world application, where moving sequences of elevators need to be planned. The sequences are due to all kinds of restrictions, like that the VIPs need to be served first. To formulate all of these restrictions, complex first order preconditions are used in the representation (Koehler & Schuster, 2000). As only a few planners could handle the full ADL representation, the domain was subdivided into the easier STRIPS and SIMPLE (conditional effects) classes, the full ADL class, and an even more expressive class where numerical constraints (the number of passengers in the elevator at a time) needed to be considered. We show the runtime curves for the participants in the full ADL class in Figure 8. In difference to the previous domains, the *Miconic* domain was run on site at AIPS-2000, using 450 MHz Pentium III machines with 256 MByte main memory.

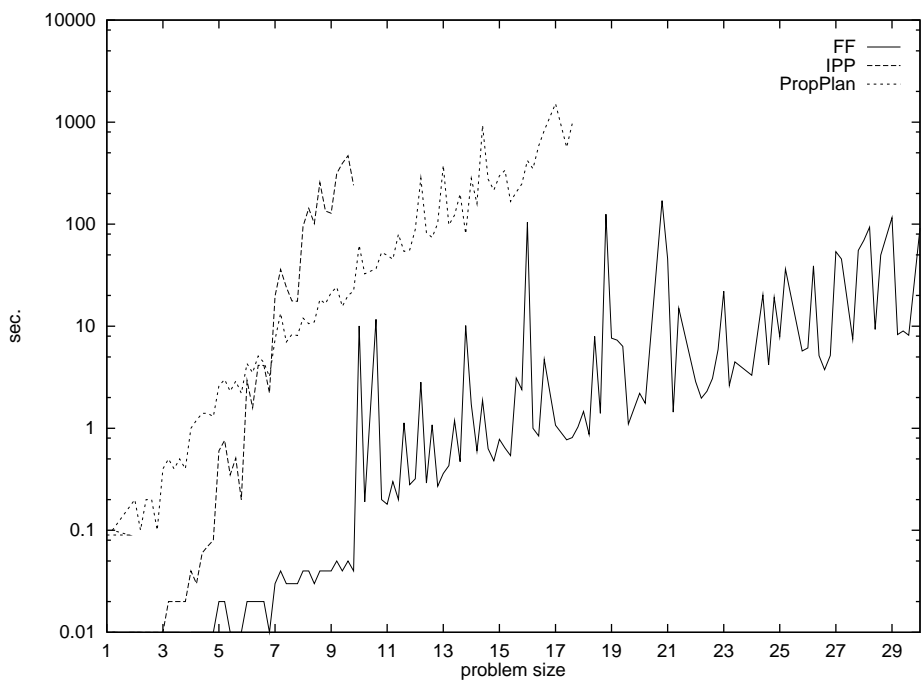

Figure 8: Runtime curves on *Elevator* tasks for those planners which handled the full ADL Miconic 10 domain representation. Time is shown on a logarithmic scale.

FF outperforms the two other full ADL planners in terms of solution time. It must be noticed, however, that IPP and PropPlan generate provably optimal plans here, such that one needs to be careful when directly comparing those running times. On the other hand, FF's plans are quite close to optimal on these instances, being within in a range of maximally 133% of the optimal solution lengths to the instances solved by PropPlan, 111% on average.

The large variation of FF's running times is apparently due to the same phenomenon as the variation in *Freecell* is: sometimes, as we observed, enforced hill-climbing runs into a dead end, which causes a switch to best-first search, solving the task in more time, but reliably. The helpful actions percentage takes very low values on average, around 15%, and breadth first search rarely goes deeper than four or five steps, where the large majority of the better successors lie at depth 1.

## 8.2 Some more Examples

In this section, we present some of the results that we have obtained in domains that were not used in the AIPS-2000 competition. We give some more examples of domains where FF works well, and, to illustrate our intuitions on the reasons for FF's behavior, also some examples of domains where FF is less appropriate.

For evaluation, we ran FF on a collection of 20 benchmark planning domains, including all domains from the AIPS-1998 and AIPS-2000 competitions, and seven more domains from the literature. Precisely, the domains in our suite were *Assembly*, two *Blocksworlds* (three- and four-operator representation), *Briefcaseworld*, *Bulldozer*, *Freecell*, *Fridge*, *Grid*, *Gripper*, *Hanoi*, *Logistics*, *Miconic*-ADL, *Miconic*-SIMPLE, *Miconic*-STRIPS, *Movie*, *Mprime*, *Mystery*, *Schedule*, *Tireworld*, and *Tsp*. Instances were either taken from published distributions, from the literature, or modified to show scaling behavior.[5] Times for FF were measured on a Sparc Ultra 10 running at 350 MHz, with a main memory of 256 MBytes. Running times that we show for other planners were taken on the same machine, if not otherwise indicated in the text. We found that FF shows extremely competitive performance on 16 of the 20 domains listed above. On the two *Blocksworlds*, *Mprime*, and *Mystery*, it still shows satisfying behavior. Some examples that have not been used in the AIPS-2000 competition are:

- The *Assembly* Domain. FF solves 25 of the 30 tasks in the AIPS-1998 test suite in less than five seconds, where the five others are either unsolvable, or have specification errors. The only other planner we know of that can solve any of the *Assembly* tasks is IPP. The latest version IPP4.0 solves only four of the very small instances, taking up to 12 hours running time. FF's plan lengths are, in terms of the number of actions, shorter than IPP's time step optimal ones, ranging from 90% to 96%.

- The *Briefcaseworld* Domain. This is a classical domain, where $n$ objects need to be transported using a briefcase. Whenever the briefcase is moved, a conditional effect forces all objects inside the briefcase to move with it. From our suite, IPP4.0 easily solves the tasks with $n \leq 5$ objects, but fails to solve any task where $n \geq 7$. FF, on the other hand, solves even the 11-objects tasks in less than a second. On the tasks that IPP solves, plan lengths of FF are within 84% to 111% of IPP's lengths, 99% on average.

- The *Grid* Domain. The 1998 competition featured five instances. For these tasks, the fastest planning mechanism we know of from the literature is a version of GRT that is enhanced with a simple kind of domain dependent knowledge, supplied by the

---

user. It solves the tasks in 1.04, 6.63, 21.35, 19.92 and 118.65 seconds on a 300 MHz Pentium Celeron machine with 64 MByte main memory (Refanidis & Vlahavas, 2000). FF solves the same tasks within 0.15, 0.47, 2.11, 1.93 and 19.54 seconds, respectively. Plan lengths of FF are within 89% to 139% of GRT's lengths, 112% on average.

- The *Gripper* Domain, used in the 1998 competition. The number of states that FF evaluates before returning an optimal sequential solution is *linear* in the size of the task there. The biggest AIPS-1998 example gets solved in 0.16 seconds.

- The *Tireworld* Domain. The original task formulated by Stuart Russel asks the planner to find out how to replace a flat tire. Koehler and Hoffmann (2000a) modified the task such that an arbitrary number of $n$ tires need to be replaced. IPP3.2, using the goal agenda technique, solves the 1, 2, and 3-tire tasks in 0.08, 0.21, and 1.33 seconds, respectively, but exhausts memory resources as soon as $n \geq 4$. FF scales to much larger tasks, taking less than a tenth of a second when $n \leq 6$, still solving the 10-tire task in 0.33 seconds. FF's plan lengths are, on the tasks that IPP manages to solve, equally long in terms of the number of actions.

As was already said, our intuition is that the majority of the currently available benchmark planning domains—at least those represented by our domain collection—are "simple" in structure, and that it is this simplicity which makes them solvable so easily by a greedy algorithm such as FF. To illustrate our intuitions, we now give data for a few domains that have a less simple structure. They are therefore challenging for FF.

### 8.2.1 The *Mystery* and *Mprime* Domains

The *Mystery* and *Mprime* domains were used in the AIPS-1998 competition. Both are variations of the *Logistics* domain, where there are additional constraints on the capacity of each vehicle, and, in particular, on the amount of fuel that is available. Both domains are closely related, the only difference being that in *Mprime*, fuel items can be transported between two locations, if one of those has more than one such item. In Figure 9, we compare FF's results on both domains to that reported by Drew McDermott for the Unpop system (McDermott, 1999).

Instances are the same for both domains in Figure 9. Results for *Unpop* have been taken by McDermott on a 300 MHz Pentium-II workstation (McDermott, 1999). A dash indicates that the task couldn't be solved by the corresponding planner.

One needs to be careful when comparing the running times in Figure 9: unlike FF, coded in C, Unpop is written in Lisp. Thus, the apparent runtime superiority of FF in Figure 9 is not significant. On the contrary, Unpop seems to solve these task collections more reliably than FF: it finds solutions to four *Mystery* and three *Mprime* instances which FF does not manage to solve. None of the planners is superior in terms of solution lengths: On *Mystery*, FF ranges within 55% to 185% of Unpop's lengths, 103% on average, on *Mprime*, FF ranges within 45% to 150%, 93% on average.

We think that FF's behavior on these two domains is due to the large amount of dead ends in the corresponding state spaces—we tried to randomize FF's search strategy, running it on the *Mystery* and *Mprime* suits. Regardless of the randomization strategy we tried, on the tasks that original FF couldn't solve search ended up being stuck in a dead end. Dead ends are a frequent phenomenon in the *Mystery* and *Mprime* domains, where, for example, an important vehicle can run out of fuel. In that sense, the tasks in these domains have a more complex structure than those in a lot of other benchmark domains, where the tasks are dead-end free. Depending more or less randomly on task structure and selected actions, FF can either solve *Mystery* and *Mprime* tasks quite fast, or fails, i.e., encounters a dead end state with enforced hill-climbing. Trying to solve the tasks with complete best-first search exhausts memory resources for larger instances.

| task | Mystery Unpop time | Mystery Unpop steps | Mystery FF time | Mystery FF steps | Mprime Unpop time | Mprime Unpop steps | Mprime FF time | Mprime FF steps |
|---|---|---|---|---|---|---|---|---|
| prob-01 | 0.3 | 5 | 0.04 | 5 | 0.4 | 5 | 0.04 | 5 |
| prob-02 | 3.3 | 8 | 0.25 | 10 | 13.5 | 8 | 0.27 | 10 |
| prob-03 | 2.1 | 4 | 0.08 | 4 | 5.9 | 4 | 0.09 | 4 |
| prob-04 | - | - | - | - | 3.9 | 9 | 0.04 | 10 |
| prob-05 | - | - | - | - | 19.2 | 17 | - | - |
| prob-06 | - | - | - | - | - | - | - | - |
| prob-07 | - | - | - | - | - | - | - | - |
| prob-08 | - | - | - | - | 52.5 | 10 | 0.40 | 10 |
| prob-09 | 3.3 | 8 | - | - | 13.5 | 8 | 0.16 | 10 |
| prob-10 | - | - | - | - | 79.0 | 19 | - | - |
| prob-11 | 1.4 | 11 | 0.05 | 9 | 2.9 | 11 | 0.06 | 9 |
| prob-12 | - | - | - | - | 8.0 | 12 | 0.20 | 10 |
| prob-13 | 370.1 | 16 | - | - | 89.3 | 15 | 0.16 | 10 |
| prob-14 | 162.1 | 18 | - | - | - | - | - | - |
| prob-15 | 17.3 | 6 | 0.98 | 8 | 14.6 | 6 | 3.39 | 8 |
| prob-16 | - | - | - | - | 25.2 | 13 | 0.28 | 7 |
| prob-17 | 13.1 | 5 | 0.70 | 4 | 4.0 | 5 | 0.92 | 4 |
| prob-18 | - | - | - | - | - | - | - | - |
| prob-19 | 11.8 | 6 | - | - | 24.7 | 6 | 0.99 | 9 |
| prob-20 | 22.5 | 7 | 0.41 | 13 | 62.8 | 17 | 3.11 | 13 |
| prob-21 | - | - | - | - | 22.1 | 11 | - | - |
| prob-22 | - | - | - | - | 135.7 | 16 | 643.19 | 23 |
| prob-23 | - | - | - | - | 55.0 | 18 | 3.09 | 14 |
| prob-24 | - | - | - | - | 24.8 | 15 | 2.7 | 9 |
| prob-25 | 0.4 | 4 | 0.02 | 4 | 0.5 | 4 | 0.02 | 4 |
| prob-26 | 6.0 | 6 | 0.85 | 7 | 16.4 | 14 | 0.16 | 10 |
| prob-27 | 3.8 | 9 | 0.05 | 5 | 2.8 | 7 | 0.78 | 5 |
| prob-28 | 1.4 | 9 | 0.01 | 7 | 1.6 | 11 | 0.08 | 5 |
| prob-29 | 0.9 | 4 | 0.06 | 4 | 1.5 | 4 | 0.30 | 4 |
| prob-30 | 20.8 | 14 | 0.23 | 11 | 17.7 | 12 | 1.86 | 11 |

Figure 9: Running times and solution length results on the AIPS-1998 *Mystery* and *Mprime* suites.

## 8.2.2 Random SAT Instances

Our last example domain is not a classical planning benchmark. To give an example of a planning task collection where FF *really* encounters difficulties, we created a planning domain containing hard random SAT instances. Figure 10 shows runtime curves for FF, IPP4.0, and BLACKBOX3.6.

The tasks in Figure 10 are solvable SAT instances that were randomly generated according to the fixed clause-length model with 4.3 times as many clauses as variables (Mitchell, Selman, & Levesque, 1992). Random instance generation and translation software to PDDL have both been provided by Jussi Rintanen. Our figure shows running times for SAT instances with 5, 10, 15, 20, 25, and 30 variables, five tasks of each size. Values for tasks of the same size are displayed in turn, i.e., all data points below 10 on the x-axis show running times for 5 variable tasks, and so on. Though the data set is small, the observation to be made is clear: FF can only solve the small instances, and two of the bigger ones. IPP and BLACKBOX scale much better, with the tendency that BLACKBOX is fastest.

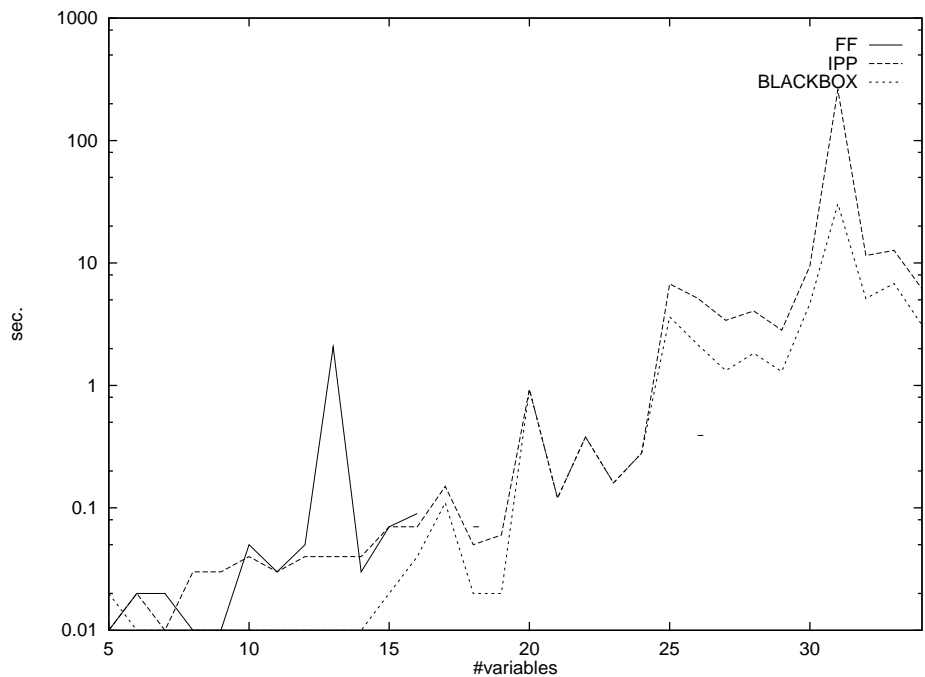

Figure 10: Runtime curves for FF, IPP, and BLACKBOX, when run on hard random SAT instances with an increasing number of variables.

The encoding of the SAT instances is the following. An operator corresponds to assigning a truth value to a variable, which makes all clauses true that contain the respective literal. Once a variable has been assigned, its value is fixed. The goal is having all clauses true. It is not surprising that BLACKBOX does best. After all, this planner uses SAT technology for solving the tasks.[6] For IPP and FF, the search space is the space of all partial truth assignments. Due to exclusion relations, IPP can rule out quite many such assignments early, when it finds they can't be completed. FF, on the other hand, does no such reasoning, and gets lost in the exponential search space, using a heuristic that merely tells it how many variables it will still need to assign truth values to, unaware of the interactions that might, and most likely will, occur.

In contrast to most of the current benchmark planning domains, finding a non-optimal solution to the planning tasks used here is NP-hard. FF's behavior on these tasks supports our intuition that FF's efficiency is due to the inherent simplicity of the planning benchmarks.

---

6. In these experiments, we ran BLACKBOX with the default parameters. Most likely, one can boost the performance by parameter tuning.

### 8.3 What Makes the Difference to HSP?

One of the questions that the authors have been asked most frequently at the AIPS-2000 planning competition is this: If FF is so closely related to HSP, then why does it perform so much better? FF uses the same basic ideas as classical HSP, forward search in state space, and heuristic evaluation by ignoring delete lists (Bonet & Geffner, 1998). The differences lie in the way FF estimates goal distances, the search strategy, and FF's pruning techniques. To obtain a picture of which new technique yields which performance results, we conducted a number of experiments where those techniques could be turned on and off independently of each other. Using all combinations of techniques, we measured runtime and solution length performance on a large set of planning benchmark tasks. In this section, we describe the experimental setup, and summarize our findings. The raw data and detailed graphical representations of the results are available in an online appendix.

#### 8.3.1 EXPERIMENTAL SETUP

We focused our investigation on FF's key features, i.e., we restricted our experiments to the FF base architecture, rather than taking into account all of FF's new techniques. Remember that FF's base architecture (cf. Section 2) is the enforced hill-climbing algorithm, using FF's goal distances estimates, and pruning the search space with the helpful actions heuristic. The additional techniques integrated deal with special cases, i.e., the added goal deletion heuristic and the goal agenda are concerned with goal orderings, and the complete best-first search serves as a kind of safety net when local search has run into a dead end. Considering all techniques independently would give us $2^6 = 64$ different planner configurations. As each of the special case techniques yields savings only in a small subset (between 4 and 6) of our 20 domains, large groups of those 64 configurations would behave exactly the same on the majority of our domains. We decided to concentrate on FF's more fundamental techniques. The differences between classical HSP and FF's base architecture are the following:

1. Goal distance estimates: while HSP approximates relaxed solution lengths by computing certain weight values, FF extracts explicit relaxed solutions, cf. Section 4.

2. Search strategy: while classical HSP employs a variation of standard hill-climbing, FF uses enforced hill-climbing as was introduced in Section 5.

3. Pruning technique: while HSP expands all children of any search node, FF expands only those children that are considered helpful, cf. Section 6.1.

We have implemented experimental code where each of these algorithmic differences is attached to a switch, turning the new technique on or off. The eight different configurations of the switches yield eight different heuristic planners. When all switches are on, the resulting planner is exactly FF's base architecture. With all switches off, our intention was to imitate classical HSP, i.e., HSP1 as it was used in the AIPS-1998 competition. Concerning the goal distance estimates switch and the pruning techniques switch, we implemented the original methods. Concerning the search strategy, we used the following simple hill-climbing design:

- Always select one best evaluated successor randomly.

- Keep a memory of past states to avoid cycles in the hill-climbing path.

- Count the number of consecutive times in which the child of a node does not improve the heuristic estimate. If that counter exceeds a threshold, then restart, where the threshold is 2 times the initial state's goal distance estimate.

- Keep visited nodes in memory across restart trials in order to avoid multiple computation of the heuristic for the same state.

In HSP1, some more variations of restart techniques are implemented. In personal communication with Blai Bonet and Hector Geffner, we decided not to imitate those variations—which affect behavior only in a few special cases—and use the simplest possible design instead. We compared the performance of our implementation with all switches turned off to the performance of HSP1, running the planners on 12 untyped STRIPS domains (the input required for HSP1). Except in four domains, the tasks solved were the same for both planners. In *Freecell* and *Logistics*, our planner solved more tasks, apparently due to implementation details: though HSP1 did not visit more states than our planner on the smaller tasks, it ran out of memory on the larger tasks. In *Tireworld* and *Hanoi*, the restarting techniques seem to make a difference: In *Tireworld*, HSP1 cannot solve tasks with more than one tire because it always restarts before getting close to the goal (our planner solves tasks with up to 3 tires), whereas in *Hanoi* our implementation can not cope with more than 5 discs for the same reason (HSP1 solves tasks with up to 7 discs). Altogether, in most cases there is a close correspondence between the behavior of HSP1 and our configuration with all switches turned off. In any case, our experiments provide useful insights into the performance of enforced hill-climbing compared to a simple straightforward hill-climbing strategy.

To obtain data, we set up a large example suite, containing a total of 939 planning tasks from our 20 benchmark domains. As said at the beginning of Section 8.2, our domains were *Assembly*, two *Blocksworld*s (three- and four-operator representation), *Briefcaseworld*, *Bulldozer*, *Freecell*, *Fridge*, *Grid*, *Gripper*, *Hanoi*, *Logistics*, *Miconic*-ADL, *Miconic*-SIMPLE, *Miconic*-STRIPS, *Movie*, *Mprime*, *Mystery*, *Schedule*, *Tireworld*, and *Tsp*. In *Hanoi*, there were 8 tasks—3 to 10 discs to be moved—in the other domains, we used from 30 to 69 different instances. As very small instances are likely to produce noisy data, we tried to avoid those by rejecting tasks that were solved by FF in less than 0.2 seconds. This was possible in all domains but *Movie*, where all tasks in the AIPS-1998 suite get solved in at most 0.03 seconds. In the two *Blocksworld* representations, we randomly generated tasks with 7 to 17 blocks, using the state generator provided by John Slaney and Sylvie Thiebaux (2001). In *Assembly* and *Grid*, we used the AIPS-1998 instances, plus a number of randomly generated ones similar in size to the biggest examples in the competition suites. In *Gripper*, our tasks contained from 10 to 59 balls to be transported. In the remaining 9 competition domains, we used the larger instances of the respective competition suites. In *Briefcaseworld* and *Bulldozer*, we randomly generated around 50 large tasks, with 10 to 20 objects, and 14 to 24 locations, respectively. In *Fridge*, from 1 to 14 compressors had to be exchanged, in *Tireworld*, 1 to 30 wheels needed to be replaced, and in *Tsp*, 10 to 59 locations needed to be visited.[7]

For each of the eight configurations of switches, we ran the respective planner on each of the tasks in our example suite. Those configurations using randomized hill-climbing were run five times on each task, and the results averaged afterwards. Though five trials might sound like a small number here—way too small if we were to compare different hill-climbing strategies for SAT problems, for example—the number seemed to be reasonable to us: remember that, in the planning framework, all hill-climbing trials start from the same state. The variance that we found between different trials was usually low in our testing runs. To complete the experiments in a reasonable time, we restricted memory consumption to 128 MByte, and time consumption to 150 seconds—usually, if FF needs more time or memory on a planning task of reasonable size, then it doesn't manage to solve it at all. As said at the beginning of the section, the raw data is available in an online appendix, accompanied by detailed graphical representations. Here, we summarize the results, and discuss the most interesting observations. We examined the data separately for each domain, as our algorithmic techniques typically show similar behavior for all tasks within a domain. In contrast, there can be essential differences in the behavior of the same technique when it is applied to tasks from different domains.

### 8.3.2 RUNNING TIME

For our running time investigation, if a configuration did not find a solution plan to a given task, we set the respective running time value to the time limit of 150 seconds (sometimes, a configuration can terminate faster without finding a plan, for example an enforced hill-climbing planner running into a dead end). In the following, we designate each switch configuration by 3 letters: "H" stands for helpful actions on, "E" stands for enforced hill-climbing on, "F" stands for FF estimates on. If a switch is turned off, the respective letter is replaced by a "−": FF's base architecture is configuration "HEF", our HSP1 imitation is "− − −", and "H−−", for example, is hill-climbing with HSP goal distances and helpful actions pruning. For a first impression of our running time results, see the averaged values per domain in Figure 11.

Figure 11 shows, for each domain and each configuration, the averaged running time over all instances in that domain. As the instances in each domain are not all the same size, but typically scale from smaller to very large tasks, averaging over all running times is, of course, a very crude approximation of runtime behavior. The data in Figure 11 provides a general impression of our runtime results per domain, and gives a few hints on the phenomena that might be present in the data. Compare, for example, the values on the right hand side—those planners using helpful actions—to those on the left hand side—those planners expanding all sons of search nodes. In *Briefcaseworld* and *Bulldozer*, the right hand side values are higher, but in almost all other domains, they are considerably lower. This is especially true for the two rightmost columns, showing values for planners using helpful actions and enforced hill-climbing. This indicates that the main sources of performance lie in the pruning technique and the search strategy—looking at the rightmost "HE−" and "HEF" columns, which only differ in the goal distance estimate, those two configuration values are usually close to each other, compared to the other configurations in the same domain.

---

7. All PDDL files, and the source code of all instance generators we used, are available in an online appendix. The generators, together with descriptions of the randomization strategies, are also available at http://www.informatik.uni-freiburg.de/˜hoffmann/ff-domains.html.

| | $---$ | $--F$ | $-E-$ | $-EF$ | $H--$ | $H-F$ | $HE-$ | $HEF$ |
|---|---|---|---|---|---|---|---|---|
| Assembly | 117.39 | 31.75 | 92.95 | 61.10 | 47.81 | 20.25 | 20.34 | 16.94 |
| Blocksworld-3ops | 4.06 | 2.53 | 8.37 | 30.11 | 1.41 | 0.83 | 0.27 | 6.11 |
| Blocksworld-4ops | 0.60 | 8.81 | 80.02 | 56.20 | 1.21 | 10.13 | 25.19 | 40.65 |
| Briefcaseworld | 16.35 | 5.84 | 66.51 | 116.24 | 150.00 | 150.00 | 150.00 | 150.00 |
| Bulldozer | 4.47 | 3.24 | 31.02 | 15.74 | 81.90 | 126.50 | 128.40 | 141.04 |
| Freecell | 65.73 | 46.05 | 54.15 | 51.27 | 57.35 | 42.68 | 43.99 | 41.44 |
| Fridge | 28.52 | 53.58 | 31.89 | 52.60 | 0.85 | 0.69 | 1.88 | 2.77 |
| Grid | 138.06 | 119.53 | 115.05 | 99.18 | 115.00 | 95.10 | 18.73 | 11.73 |
| Gripper | 2.75 | 1.21 | 15.16 | 1.00 | 1.17 | 0.48 | 0.17 | 0.11 |
| Hanoi | 93.76 | 75.05 | 6.29 | 3.91 | 150.00 | 78.82 | 4.47 | 2.70 |
| Logistics | 79.27 | 102.09 | 79.77 | 111.47 | 36.88 | 39.69 | 10.18 | 11.94 |
| Miconic-ADL | 150.00 | 150.00 | 102.54 | 54.23 | 142.51 | 128.28 | 95.45 | 59.00 |
| Miconic-SIMPLE | 2.61 | 2.01 | 2.47 | 1.93 | 1.35 | 0.86 | 0.55 | 0.56 |
| Miconic-STRIPS | 2.71 | 2.32 | 4.84 | 1.53 | 1.44 | 1.01 | 0.64 | 0.36 |
| Movie | 0.02 | 0.02 | 0.02 | 0.02 | 0.02 | 0.02 | 0.02 | 0.02 |
| Mprime | 73.09 | 69.27 | 82.89 | 81.43 | 47.09 | 58.45 | 18.56 | 26.62 |
| Mystery | 78.54 | 90.55 | 71.60 | 86.01 | 75.73 | 95.24 | 85.13 | 86.21 |
| Schedule | 135.50 | 131.12 | 143.59 | 141.42 | 77.58 | 38.23 | 12.23 | 13.77 |
| Tireworld | 135.30 | 110.38 | 119.22 | 121.34 | 121.13 | 105.67 | 97.41 | 85.64 |
| Tsp | 4.11 | 0.82 | 2.45 | 0.75 | 2.48 | 0.57 | 0.15 | 0.07 |

Figure 11: Averaged running time per domain for all eight configurations of switches.

To put our observations on a solid basis, we looked, for each domain, at each pair of configurations in turn, amounting to $20 * \frac{8*7}{2} = 560$ pairs of planner performances. For each such pair, we decided whether one configuration performed significantly better than the other one. To decide significance, we counted the number of tasks that one configuration solved faster. We found this to be a more reliable criterion than things like the difference between running times for each task. As tasks grow in size, rather than being taken from a population with finite mean size, parametric statistical procedures, like computing confidence intervals for runtime differences, make questionable assumptions about the distribution of data. We thus used the following non-parametric statistical test, known as the two-tailed sign test (Siegel & N. J. Castellan, 1988). Assume that both planners, A and B, perform equally on a given domain. Then, given a random instance from the domain, the probability that B is faster than A should be equal to the probability that A is faster than B. Take this as the null hypothesis. Under that hypothesis, if A and B behave differently on an instance, then B is faster than A with probability $\frac{1}{2}$. Thus, the tasks where B is faster are distributed over the tasks with different behavior according to a Binomial distribution with $p = \frac{1}{2}$. Compute the probability of the observed outcome under the null hypothesis, i.e., if there are $n$ tasks where A and B behave differently, and $k$ tasks where B is faster, then compute the probability that, according to a binomial distribution with $p = \frac{1}{2}$, at least $k$ positive outcomes are obtained in $n$ trials. If that probability is less or equal than .01, then reject the null hypothesis and say that B performs significantly better than A. Symmetrically, decide whether A performs significantly better than B. We remark that in all domains except *Movie* the tasks where two configurations behaved equally were exactly those that could not be solved by either of the configurations. In 60% of the cases where we found that one configuration B performed significantly better than another configuration, B was faster on *all* instances with different behavior. In 71%, B was faster on all but one such instance.

We are particularly interested in pairs A and B of configurations where B results from A by turning one of the switches on, leaving the two others unchanged. Deciding about significant improvement in such cases tells us about the effect that the respective technique has on performance in a domain. There are 12 pairs of configurations where one switch is turned on. Figure 12 shows our findings in these cases.

| domain | F | | | | E | | | | H | | | |
|---|---|---|---|---|---|---|---|---|---|---|---|---|
| | —— | –E | H– | HE | —— | –F | H– | HF | —— | –F | E– | EF |
| Assembly | + | + | + | + | | – | + | + | + | + | + | + |
| Blocksworld-3ops | + | | + | – | + | – | + | + | + | + | + | + |
| Blocksworld-4ops | | + | | | – | – | – | | | – | + | + |
| Briefcaseworld | + | – | | | | – | – | | | – | – | – |
| Bulldozer | + | + | | | | – | – | – | | – | – | – |
| Freecell | + | + | + | + | + | + | + | + | + | + | + | + |
| Fridge | – | | | | | – | – | – | + | + | + | + |
| Grid | + | | + | + | + | | + | + | + | + | + | + |
| Gripper | + | + | + | + | – | + | + | + | + | + | + | + |
| Hanoi | | + | | | + | + | + | + | | | | |
| Logistics | – | – | + | | | – | + | + | + | + | + | + |
| Miconic-ADL | | + | | + | + | + | + | + | | + | + | + |
| Miconic-SIMPLE | + | + | + | | + | + | + | + | + | + | + | + |
| Miconic-STRIPS | + | + | + | + | | + | + | + | + | + | + | + |
| Movie | | | | | + | + | + | + | | | | |
| Mprime | + | + | + | | | + | + | + | + | + | + | + |
| Mystery | | | | | + | | + | + | + | | | |
| Schedule | | | + | + | | – | + | + | + | + | + | + |
| Tireworld | + | | + | + | | – | + | + | | + | + | + |
| Tsp | + | + | + | + | + | + | + | + | + | + | + | + |

Figure 12: The effect of turning on a single switch, keeping the others unchanged. Summarized in terms of significantly improved or degraded running time performance per domain, and per switch configuration.

Figure 12 is to be understood as follows. It shows our results for the "F", "E", and "H" switches, which become active in turn from left to right. For each of these switches, there are four configurations of the two other, background, switches, displayed by four columns in the table. In each column, the behavior of the respective background configuration with the active switch turned off is compared to the behavior with the active switch turned on. If performance is improved significantly, the table shows a "+", if it is significantly degraded, the table shows a "−", and otherwise the respective table entry is empty. For example, consider the top left corner, where the "F" switch is active, and the background configuration is "−−", i.e., hill-climbing without helpful actions. Planner A is "− − −", using HSP distances, and planner B is "−−F", using FF distances. B's performance is significantly better than A's, indicated by a "+".

The leftmost four columns in Figure 12 show our results for HSP distance estimates versus FF distance estimates. Clearly, the latter estimates are superior in our domains, in the sense that, for each background configuration, the behavior gets significantly improved in 8 to 10 domains. In contrast, there are only 5 cases altogether where performance gets worse. The significances are quite scattered over the domains and background configurations, indicating that a lot of the significances result from interactions between the techniques that occur only in the context of certain domains. For example, performance is improved in *Bulldozer* when the background configuration does not use helpful actions, but degraded when the background configuration uses hill-climbing with helpful actions. This kind of behavior can not be observed in any other domain. There are 4 domains where performance is improved in all but one background configuration. Apparently in these cases some interaction between the techniques occurs only in one specific configuration. We remark that often running times with FF's estimates are only a little better than with HSP's estimates, i.e., behavior gets improved reliably over all instances, but only by a small factor (to get an idea of that, compare the differences between average running times in Figure 11, for configurations where only the distance estimate changes). In 5 domains, FF's estimates improve performance consistently over all background configurations, indicating a real advantage of the different distance estimates. In *Gripper* (described in Section 6.1), for example, we found the following. If the robot is in room A, and holds only one ball, FF's heuristic prefers picking up another ball over moving to room B, i.e., the picking action leads to a state with better evaluation. Now, if there are $n$ balls left in room A, then HSP's heuristic estimate of picking up another ball is $4n - 2$, while the estimate of moving to room B is $3n + 1$. Thus, if there are at least 4 balls left in room A, moving to room B gets a better evaluation. Summing up weights, HSP overestimates the usefulness of the moving action.

Comparing hill-climbing versus enforced hill-climbing, i.e., looking at the four columns in the middle of Figure 12, the observation is this. The different search technique is a bit questionable when the background configuration does not use helpful actions, but otherwise, enforced hill-climbing yields excellent results. Without helpful actions, performance gets degraded almost as many times as it gets improved, whereas, with helpful actions, enforced hill-climbing improves performance significantly in 16 of our 20 domains, being degraded only in *Fridge*. We draw two conclusions. First, whether one or the other search strategy is adequate depends very much on the domain. A simple example for that is the *Hanoi* domain, where hill-climbing always restarts before it can reach the goal—on all paths to the goal, there are exponentially many state transitions where the son has no better evaluation than the father. Second, there is an interaction between enforced hill-climbing and helpful actions pruning that occurs consistently across almost all of our planning domains. This can be explained by the effect that the pruning technique has on the different search strategies.

In hill-climbing, helpful actions pruning prevents the planner from looking at too many superfluous successors on each single state that a path goes through. This saves time proportional to the length of the path. The effects on enforced hill-climbing are much more drastic. There, helpful actions prunes out unnecessary successors of each state during a breadth first search, i.e., it cuts down the branching factor, yielding performance speedups exponential in the depths that are encountered.

We finally compare consideration of all actions versus consideration of only the helpful ones. Look at the rightmost four columns of Figure 12. The observation is simply that helpful actions are really helpful—they improve performance significantly in almost all of our planning domains. This is especially true for those background configurations using enforced hill-climbing, due to the same interaction that we have outlined above. In some domains, helpful actions pruning imposes a very rigid restriction on the search space: in *Schedule*, as said in Section 8.1.3, we found that states can have hundreds of successors, where only about 2% of those are considered helpful. In other domains, only a few actions are pruned, like in *Hanoi*, where at most three actions are applicable in each state, which are all considered helpful in most of the cases. Even a small degree of restriction does usually lead to a significant improvement in performance. In two domains, *Briefcaseworld* and *Bulldozer*, helpful actions can prune out too many possibilities, i.e., they cut away solution paths. This happens because there, the relaxed plan can ignore things that are crucial for solving the real task. Consider the *Briefcaseworld*, briefly described in Section 8.2, where objects need to be moved using a briefcase. Whenever the briefcase is moved, all objects inside it are moved with it by a conditional effect. Now, the relaxed planner never needs to take any object out of the briefcase—the delete effects say that moving an object means the object is no longer at the start location. Ignoring this, keeping objects inside the briefcase never hurts.

### 8.3.3 SOLUTION LENGTH

We also investigated the effects that FF's new techniques have on solution length. Comparing two configurations A and B, we took as the data set the respective solution length for those tasks that both A and B managed to solve—obviously, there is not much point in comparing solution length when one planner can not find a solution at all. We then counted the number $n$ of tasks where A and B behaved differently, and the number $k$ where B's solution was shorter, and decided about significance like described in the last section. Figure 13 shows our results in those cases where a single switch is turned.

The data in Figure 13 are organized in the obvious manner analogous to Figure 12. A first glance at the table tells us that FF's new techniques are also useful for shortening solution length in comparison to HSP1, but not as useful as they are for improving runtime behavior. Let us focus on the leftmost four columns, HSP distance estimates versus FF distance estimates. The observations are that, with enforced hill-climbing in the background, FF estimates often result in shorter plans, and that there are two domains where solution lengths are improved across all background configurations. Concerning the second observation, this is due to properties of the domain that FF's heuristic recognizes, but HSP's doesn't. Recall what we observed about the *Gripper* domain in the preceding section. With the robot standing in room A, holding only one ball, the FF heuristic gives picking up the ball a better evaluation than moving to room B. The HSP heuristic doesn't do this. Therefore, using the HSP heuristic results in longer plans. Concerning the first observation, improved solution lengths when enforced hill-climbing is in the background, we do not have a good explanation for this. It seems that the greedy way in which enforced hill-climbing builds its plans is just better suited when distance estimates are cautious, i.e., low.

| domain | F | | | | E | | | | H | | | |
|---|---|---|---|---|---|---|---|---|---|---|---|---|
| | −− | −E | H− | HE | −− | −F | H− | HF | −− | −F | E− | EF |
| Assembly | | + | + | + | | + | + | | + | + | | |
| Blocksworld-3ops | | + | | + | | + | | + | | + | + | |
| Blocksworld-4ops | | | | + | + | + | + | + | + | + | + | |
| Briefcaseworld | | + | | | + | + | | | | | | |
| Bulldozer | | | | | + | + | | | | | | |
| Freecell | | + | | + | | | | | + | + | | |
| Fridge | − | | − | | + | + | | + | + | + | | |
| Grid | | | | + | + | + | + | + | + | + | | |
| Gripper | + | + | + | + | + | | − | | | | − | |
| Hanoi | | | | | | | | | | | | |
| Logistics | | + | − | + | + | + | + | + | + | | + | |
| Miconic-ADL | | + | | + | | | | + | | | | + |
| Miconic-SIMPLE | − | + | + | + | | + | + | + | | + | + | + |
| Miconic-STRIPS | + | + | + | + | + | + | + | + | | | + | + |
| Movie | | | | | − | − | − | − | | | | |
| Mprime | | | | | | | | | | | | |
| Mystery | | | | | | | | | | | | |
| Schedule | | | + | | | | | − | + | + | | |
| Tireworld | | | | + | | | | + | | | | |
| Tsp | | | | | | | | | | | | |

Figure 13: The effect of turning on a single switch, keeping the others unchanged. Summarized in terms of significantly improved or degraded solution length performance per domain, and per switch configuration.

Consider the four columns in the middle of Figure 13, hill-climbing versus enforced hill-climbing. There are many cases where the different search strategy results in shorter plans. We figure that this is due to the different plateau behavior that the search methods exhibit, i.e., their behavior in flat regions of the search space. Enforced hill-climbing enters a plateau somewhere, performs complete search for a state with better evaluation, and adds the shortest path to that state to its current plan prefix. When hill-climbing enters a plateau, it strolls around more or less randomly, until it hits a state with better evaluation, or has enough of it and restarts. All the actions on its journey to the better state are kept in the final plan. In *Movie*, the phenomenon is this. If a planner chooses to reset the counter on the VCR before it chooses to rewind the movie (initially, neither heuristic makes a distinction between these two actions), then it has to reset the counter again. The enforced hill-climbing planners always reset the counter first. The hill-climbing planners, on the other hand, randomly choose either ordering with equal probability. As said in Section 8.3.1, hill-climbing was given five tries on each task, and results averaged. In five tries, around half of the solutions use the correct ordering, such that, for all tasks, the average value is lower than the corresponding value for the enforced hill-climbing planners.

Finally, we compare consideration of all actions versus consideration of only the helpful ones, results depicted in the rightmost four columns of Figure 12. Coming a bit unexpected, there is only one single case where solution length performance is degraded by turning on helpful actions. This indicates that the actions on the shortest path to the goal are, in fact, usually considered helpful—unless *all* solution paths are thrown away, as is sometimes the case only in the *Briefcaseworld* and *Bulldozer* domains. Quite the other way around than one should think, pruning the search space with helpful actions sometimes leads to significantly shorter solution plans, especially when the underlying search method is hill-climbing. Though this may sound paradoxical, there is a simple explanation to it. Consider what we said above about the plateau behavior of hill-climbing, randomly adding actions to the current plan in the search for a better state. If such a search engine is armed with the helpful actions successors choice, focusing it into the direction of the goals, it might well take less steps to find the way off a plateau.

## 9. Related Work

The most important connections of the FF approach to methodologies reported in the literature are the following:

- HSP's basic idea of forward state space search and heuristic evaluation by ignoring delete lists (Bonet & Geffner, 1998).

- The view of our heuristic as a special case of GRAPHPLAN (Blum & Furst, 1995), and its connection to HSP's heuristic method.

- The similarity of the helpful actions heuristic to McDermott's favored actions (1996), and to irrelevance detection mechanisms (Nebel et al., 1997).

- The inspiration of the added goal deletion heuristic by work done by Koehler and Hoffmann (2000a), and the adaption of the goal agenda approach (Koehler, 1998).

- The adaption of IPP's ADL preprocessing phase (Koehler & Hoffmann, 2000b), inspired by ideas from Gazen and Knoblock (1997).

We have discussed all of these connections in the respective sections already. So let us focus on a connection that has not yet been mentioned. It has been recognized after the first planning competition at AIPS-1998 that the main bottleneck in HSP1 is the recomputation of the heuristic on each single search state. Two recent approaches are based on the observation that the repeated recomputation is necessary because HSP1 does forward search with a forward heuristic, i.e., the directions of search and heuristic are the same. The authors of HSP themselves stick to their heuristic, but change the search direction, going backwards from the goal in HSP-r (Bonet & Geffner, 1999). This way, they need to compute weight values only once, estimating each fact's distance to the initial state, and only sum the weights up for a state later during search.[8]

Refanidis and Vlahavas (1999) invert the direction of the HSP heuristic instead. While HSP computes distances going from the current state towards the goal, GRT goes from the goal to each fact. The function that then extracts, for each state during forward search, the states heuristic estimate, uses the pre computed distances as well as some information on which facts will probably be achieved simultaneously.

Interestingly, FF recomputes, like HSP, the heuristic from scratch on each search state, but nevertheless outperforms the other approaches. As we have seen in Section 8.3, this is for the most part due to FF's search strategy and the helpful actions pruning technique.

## 10. Conclusion and Outlook

We have presented an approach to domain independent planning that, at the time being, outperforms all existing technology on the majority of the currently available benchmark domains. Just like the well known HSP1 system, it relies completely on forward state space search and heuristic evaluation of states by ignoring delete lists. Unlike HSP, the method uses a GRAPHPLAN-style algorithm to find an explicit relaxed solution to each search state. Those solutions give a more careful estimation of a state's difficulty. As a second major difference to HSP, our system employs a novel local search strategy, combining hill-climbing with complete search. Finally, the method makes use of powerful heuristic pruning techniques, which are based on examining relaxed solutions.

As we have mentioned earlier, our intuition is that the reasons for FF's efficiency lie in structural properties that the current planning benchmarks tend to have. As a matter of fact, the simplicity of the benchmarks quite immediately meets the eye, once one tries to look for it. It should be clear that the *Gripper* tasks, where some balls need to be transported from one room to another, exhibit a totally different search space structure than, for example, hard random SAT instances. Therefore, it's intuitively unsurprising that different search methods are appropriate for the former tasks than are traditionally used for the latter. The efficiency of FF on many of the benchmarks can be seen as putting that observation to the surface.

To make explicit the hypotheses stated above, we have investigated the state spaces of the planning benchmarks. Following Frank et al. (1997), we have collected empirical data, identifying characteristic parameters for different kinds of planning tasks, like the density and size of local minima and plateaus in the search space. This has lead us to a taxonomy for planning domains, dividing them by the degree of complexity that the respective task's state spaces exhibit with respect to relaxed goal distances. Most of the current benchmark domains apparently belong to the "simpler" parts of that taxonomy (Hoffmann, 2001). We also approach our hypotheses from a theoretical point of view, where we measure the degree of interaction that facts in a planning task exhibit, and draw conclusions on the search space structure from that. Our goal in that research is to devise a method that automatically decides which part of the taxonomy a given planning task belongs to.

In that context, there are some remarks to be made on what AI planning research is heading for. Our point of view is that the goal in the field should not be to develop a technology that works well on *all* kinds of tasks one can express with planning languages. This will hardly be possible, as even simple languages as STRIPS can express NP-hard problems like SAT. What might be possible, however, is to devise a technology that works well on those tasks that *can* be solved efficiently. In particular, if a planning task does not constitute much of a problem to an uninformed human solver, then it neither should do so to our planning algorithms. With the FF system, we already seem to have a method that accomplishes this quite well, at least for sequential planning in STRIPS and ADL. While FF is not particularly well suited for solving random SAT instances, it easily solves intuitively simple tasks like the *Gripper* and *Logistics* ones, and is well suited for a number of other domains where finding a non-optimal solution is not NP-hard. This sheds a critical light on the predictions of Kautz and Selman (1999), who suspected that planning technology will become superfluous because of the fast advance of the state of the art in propositional reasoning systems. The methods developed there are surely useful for solving SAT. They might, however, not be appropriate for the typical structures of tasks that AI planning should be interested in.

---

8. HSP-r is integrated into HSP2 as an option of configuring the search process (Bonet & Geffner, 2001).

## Acknowledgments

This article is an extended and revised version of a paper (Hoffmann, 2000) that has been published at ISMIS-00. The authors wish to thank Blai Bonet and Hector Geffner for their help in setting up the experiments on the comparison of FF with HSP. We thank Jussi Rintanen for providing us with software to create random SAT instances in the PDDL language, and acknowledge the anonymous reviewer's comments, which helped improve the paper.